\documentclass[conference]{IEEEtran}
\IEEEoverridecommandlockouts
\def\ps@headings{%
\def\@oddhead{\mbox{}\scriptsize\rightmark \hfil \thepage}%
\def\@evenhead{\scriptsize\thepage \hfil \leftmark\mbox{}}%
\def\@oddfoot{}%
\def\@evenfoot{}} 
\makeatother 
\usepackage{graphicx}
\usepackage{epstopdf}
\usepackage{cite}
\usepackage{times}
\usepackage{dsfont} 
\usepackage{times}
\usepackage{pifont}
\usepackage{multirow}   
\usepackage{tikz}
\usetikzlibrary{shapes.geometric,calc}
   
\usepackage{amsthm}
\usepackage{graphicx}
\usepackage{subfigure}
\usepackage{color}
\usepackage{ifpdf}
\usepackage{epsfig}
\usepackage{latexsym}
\usepackage{amsfonts}
\usepackage{amssymb}
\usepackage{paralist}
\usepackage{comment}
\usepackage{xspace}
\usepackage{mathrsfs}
\usepackage{amssymb}
\usepackage{diagbox}
\usepackage{setspace}
\usepackage{color}
\usepackage[small]{caption}
\usepackage{mathtools}

\usepackage{subeqnarray}
\usepackage{algorithm}
\usepackage{algpseudocode}
\usepackage{amssymb}
\usepackage{url,epsfig,array}
\usepackage{leftidx} 
\usepackage{amsmath}
\usepackage[T1]{fontenc}
\usepackage{aecompl}

\usepackage{amsmath,amssymb,amsfonts}
\usepackage{graphicx}
\usepackage{textcomp}
\usepackage{xcolor}

\usepackage{graphicx}

 \usepackage[
top=0.7in, bottom=0.7in,
left=0.65in, right=0.65in
]{geometry}

\def\ie{\textit{i.e.}\xspace}
\def\etal{\textit{et al.}\xspace}

\def\eg{\textit{e.g.}\xspace}

\makeatletter
\renewcommand{\maketag@@@}[1]{\hbox{\m@th\normalsize\normalfont#1}}%
\makeatother

\def\BibTeX{{\rm B\kern-.05em{\sc i\kern-.025em b}\kern-.08em
    T\kern-.1667em\lower.7ex\hbox{E}\kern-.125emX}}
\begin{document} 
  
\title{Enhancing Federated Graph Learning via Adaptive Fusion of Structural and Node Characteristics}


\author{ \IEEEauthorblockN{Xianjun Gao, Jianchun Liu, Hongli Xu, Shilong Wang, Liusheng Huang  }
 \IEEEauthorblockA{
 School of Computer Science and Technology, University of Science and Technology of China\\
 Suzhou Institute for Advanced Research, University of Science and Technology of China\\}
 }

\pagestyle{plain}
\maketitle

\begin{abstract}
Federated Graph Learning (FGL) has demonstrated the advantage of training a global Graph Neural Network (GNN) model across distributed clients using their local graph data. 
Unlike Euclidean data (\eg, images), graph data is composed of nodes and edges, where the overall node-edge connections determine the topological structure, and individual nodes along with their neighbors capture local node features. 
However, existing studies tend to prioritize one aspect over the other, leading to an incomplete understanding of the data and the potential misidentification of key characteristics across varying graph scenarios.
Additionally, the non-independent and identically distributed (non-IID) nature of graph data makes the extraction of these two data characteristics even more challenging.
To address the above issues, we propose a novel FGL framework, named FedGCF, which aims to simultaneously extract and fuse structural properties and node features to effectively handle diverse graph scenarios.
FedGCF first clusters clients by structural similarity, performing model aggregation within each cluster to form the shared structural model. 
Next, FedGCF selects the clients with common node features and aggregates their models to generate a common node model. This model is then propagated to all clients, allowing common node features to be shared.
By combining these two models with a proper ratio, FedGCF can achieve a comprehensive understanding of the graph data and deliver better performance, even under non-IID distributions.
To efficiently cope with various graph scenarios, we develop an online learning algorithm that adjusts the combination ratio based on the system environment and training status, maximizing the utility of both characteristics.
Experimental results show that FedGCF improves accuracy by 4.94\%-7.24\% under different data distributions and reduces communication cost by 64.18\%-81.25\% to reach the same accuracy compared to baselines.

\end{abstract}
  
\begin{IEEEkeywords}
\emph{Federated Graph Learning, Non-IID Graph Data, Structural Properties, Node Features}.
\end{IEEEkeywords}


\section{Introduction}\label{sec:intro}
As the interconnection and interaction of numerous distributed devices continue to grow, the volume of graph data (\eg, social networks) generated in real-world applications has increased significantly \cite{ying2018graph,mislove2007measurement,hamilton2017representation}. 
However, growing concerns over privacy risks have made individuals and enterprises increasingly reluctant to share their private graph data.
As a result, it is infeasible to transfer the local graph data from distributed devices to a central server for unified utilization, which hinders the traditional centralized approach to training efficient Graph Neural Network (GNN) \cite{zhang2019heterogeneous,wu2019simplifying,li2019deepgcns}. 
To effectively utilize graph data while preserving privacy, Federated Graph Learning (FGL), as a new distributed machine learning paradigm, has been proposed to allow massive devices (or clients) to collaboratively train a global GNN model without sharing their private local graph data \cite{li2022federated,yao2024fedgcn}. A typical FGL system comprises a parameter server (PS) and multiple clients \cite{yao2024fedgcn,zhang2023glasu,wu2021fedgnn}. The PS distributes a global GNN model to the clients, allowing them to train the model on their local graph data. After local training, the clients upload the updated model parameters back to the PS for the global model aggregation. In this way, the global model can effectively learn from each client's data while maintaining personal data privacy and ensuring compliance with data protection regulations.

Different from the Euclidean data (\eg, image data) used in traditional federated learning (FL) \cite{wang2022enhancing,jiang2022fedmp,liu2020federated}, FGL primarily focuses on graph data, which consists of many nodes and edges. 
The global connection pattern of nodes and edges determines the overall properties of the graph data, reflecting its topological structure, while the unique information of each node and its connections with neighboring nodes depict the local features of the data \cite{yan2024peaches,newman2003structure}.
Accordingly, utilizing graph data effectively always necessitates consideration of two different types of characteristics, \ie, \textit{structural properties} and \textit{node features}. 
For structural properties, consider a small molecule classification task: benzene-like substances have a ring structure of six carbon atoms, while alkanes form a chain structure \cite{clayden2012organic}. 
Therefore, classifying these substances focuses more on determining whether their structure is ring-like or chain-like. 
Regarding node features in a social network, where nodes represent users and edges denote the social relationships, the classification of a user relationship primarily relies on the features of the connected users and their mutual friends, while non-neighboring nodes may introduce errors due to their lack of relevance \cite{marin2011social}. Thus, the classification process relies heavily on the features of the user and its connected neighboring users, with less emphasis on the broader network structure. 


Since structural properties and node features offer distinct perspectives on graph data, focusing solely on one aspect during training often leads to an incomplete understanding of the data and potential misidentification of key characteristics, resulting in significantly diminished training performance (\eg, accuracy and convergence rates).
For example, in an online shopping recommendation system, relying only on neighboring node features may limit product recommendations to highly active users, while focusing solely on structural properties will overlook individual preferences, resulting in inaccurate suggestions \cite{gao2024towards}.
Additionally, the emphasis on data characteristics may vary depending on the task objectives. For instance, in social networks, the user group classification relies more on structural properties, while analyzing relationships between users focuses on node features \cite{marin2011social}. Prioritizing structural properties in relationship analysis will cause irrelevant users to affect the classification, leading to inaccurate outcomes.

In addition to the unique data characteristics, implementing an effective FGL framework encounters another challenge of data skew, \ie, heterogeneous data distribution. 
Due to differences in client behavior and data collection methods, the data distribution on clients is often non-independent and identically distributed (non-IID) \cite{zhao2018federated,wu2023non,zhang2022fine}. 
For example, in social networks, younger users tend to form high-density graphs due to frequent interactions, while older users exhibit sparser connections, resulting in low-density graphs.
Generally, training models on the non-IID data will reduce the training efficiency and decrease test accuracy (up to 55\% \cite{li2019convergence}).
To mitigate the non-IID issue in graph data, some studies leverage the inherent graph characteristics of these data. 
Xie \etal \cite{xie2021federated} introduce GCFL, which clusters clients based on GNN gradients to reduce the impact of data heterogeneity. 
GCFL emphasizes node features but overlooks structural properties, thus reducing its effectiveness in tasks like molecule classification. 
Tan \etal \cite{tan2023federated} present FedStar, which shares structural information and captures insights through structure embeddings, excelling in molecular structure analysis. 
However, FedStar struggles with data that heavily relies on node features (\eg, publication networks). 
In summary, these methods primarily tackle non-IID challenges in graph scenarios with specific emphasis on particular characteristics, limiting their generalizability. 



In this paper, we aim to address the non-IID challenge effectively in various graph scenarios by simultaneously extracting and utilizing both structural properties and node features. 
However, this approach still introduces additional issues.
First, structural properties and node features offer different perspectives on graph data and often exist in distinct dimensions, making it difficult to extract both characteristics from different dimensions within the graph dataset. Second, the inconsistency in graph data distribution among clients leads to substantial variations in graph structure, complicating the utilization of shared structural properties. Additionally, the differences in node relationships and connections across clients further hinder the application of common node features.
These challenges significantly increase the difficulty of effectively leveraging structural properties and node features in non-IID graph data.



In order to address the aforementioned challenges, we propose an efficient FGL framework, named FedGCF, which aims to extract and fuse both structural properties and node features to handle various graph scenarios.
Specifically, FedGCF preprocesses graph data on each client using the graph traversal algorithm to encode structural properties, followed by a GNN model to extract these properties. 
Next, FedGCF clusters clients by structure similarity and aggregates models within each cluster to obtain a shared structural model.
Additionally, FedGCF constructs a connected topology based on model similarity, where vertices represent clients and the distance between vertices indicates data similarity. 
FedGCF selects clients with common node features based on this topology and aggregates their models to generate a common node model that can propagate features across all clients.
After extracting these two data characteristics, FedGCF combines the shared structural model and the common node model with a proper ratio to derive the characteristic fusion model, enabling more comprehensive learning and effective utilization of graph data characteristics, thereby enhancing model training performance.
In FedGCF, sharing structural properties among clients with similar structures fosters collaboration, reducing disparities and mitigating non-IID effects.
Shared node features ensure each client captures the most relevant features, benefiting from collective knowledge.
Nevertheless, different graph scenarios emphasize distinct characteristics, and an inappropriate combination ratio will negatively impact test accuracy and slow convergence.
For instance, in molecular classification, overemphasizing node features while overlooking structural properties will result in misinterpreted molecular functions and incorrect classifications. 
In social network user relationship prediction, relying too heavily on overall structure can introduce irrelevant information, reducing accuracy.
In addition, directly accessing data or estimating the optimal combination ratio based on historical experience is not necessarily accurate or reliable.
Therefore, \textit{how to determine the optimal combination ratio of these two characteristics for different graph scenarios remains a key challenge in FedGCF.}
The main contributions of this work are summarized as follows:
\begin{itemize}
    \item We propose FedGCF, an efficient FGL framework that simultaneously extracts and fuses structural properties and node features from graph data. 
    FedGCF fuses these two types of characteristics with a proper ratio to improve the model training performance.
    \item We design a learning-driven algorithm to adaptively adjust the characteristics combination ratio based on the current system environment and training process, enhancing the generalization on varying graph scenarios.
    \item Experimental results demonstrate that FedGCF leads to an average improvement of 7.10\% in model test accuracy. When the same target accuracy is achieved, FedGCF reduces communication cost by 64.18\%-81.25\%. Under varying data distributions, FedGCF achieves an improvement in model test accuracy of 4.94\% to 7.24\%.
\end{itemize}

\section{Preliminaries and Motivation}\label{sec:prelim}

\subsection{Federated Graph Learning (FGL)}\label{subsec:FedGNN} 
The typical FGL system always involves a parameter server (PS) and a set of $N$ clients, which enables them to collaboratively train a GNN model using local graph data from the clients \cite{gao2024towards,huang2024federated}. In FGL, instead of sharing local graph data from each client, only the GNN model parameters are exchanged between the PS and clients. This approach efficiently utilizes graph data while safeguarding the privacy of clients. Generally, each client $i$ holds local graph dataset $D_i \coloneqq (G_i, Y_i)$, where $G_i = (V_i,E_i)$ represents the graph in $D_i$ with node set $V_i$ and edge set $E_i$. $Y_i$ is the label set of $G_i$. Two nodes are connected by an edge, with each node $u$ and edge $e$ connecting nodes $u$ and $v$ having independent feature information $x_u$ and $e_{u,v}$. Therefore, compared to traditional Euclidean data such as images, graph data has a more complex structure, and the connectivity of the nodes within the data is inconsistent. Let $n_i$ represent the sample size of the dataset $D_i$ on client $i$, and $n$ denote the total number of data samples across all clients, \ie, $n=\sum^{N}_{i=1} n_i$. The complete FGL training process typically involves a total of $T$ rounds. For each round $t \in \{1, ..., T\}$, there are four steps as follows:

(1) \textbf{Global Model Distribution}: At the beginning of each round $t$, the PS distributes the global GNN model $\omega_t$ to clients. This ensures that each client begins its local model training with the most up-to-date version of the global model, maintaining consistency and improving the overall performance of the distributed learning process.

(2) \textbf{Local Model Training}: Once receiving the global model $\omega_t$, each client will perform several iterations of local GNN model training, with each iteration consisting of three main phases, \ie, neighborhood aggregation, message passing and readout \cite{he2021fedgraphnn}. In neighborhood aggregation and message passing phases, each node in the graph data needs to iteratively gather information propagated by its neighboring nodes to form the aggregated message, and then update its own node expression based on the aggregated message. To facilitate understanding, we take an $L$-layer GNN model as an example. Let $a^{l+1}_u$ represent the aggregated message and $h^l_u$ represent the state for node $u$, where $l \in [L]$ represents the layer index. The neighborhood aggregation and message passing phases can be formalized as follows:

\begin{equation}
	a^{l+1}_{u} = AGG (\{ h^l_v : v \in \mathcal{N}_u \}), \forall l \in [L]
\end{equation}
\begin{equation}
	h^{l+1}_{u} = U (h^{l}_{u}, a^{l+1}_{u})
\end{equation}
where the initial state $h^{0}_{u}$ is defined as $h^{0}_{u} = x_u$, representing the node feature, and $\mathcal{N}_u$ is the neighbor set of node $u$. $AGG(\cdot)$ serves as an aggregation function to gather information from neighboring nodes, with various GNN models applying different aggregation methods. For example, \textit{sum} is used in GCN \cite{kipf2016semi}, while \textit{mean} is used in GraphSAGE \cite{hamilton2017inductive}. $U(\cdot)$ is the state-update function, which can update the state of each node in the next layer. In the readout phase, the model leverages the node states from the final layer to make predictions for different tasks, as outlined below:

\begin{equation}
    r = readout(\{h^{L}_u \mid u \in V_i\})
\end{equation}
where $V_i$ is the node set in the graph $G_i$ and $readout(\cdot)$ aggregates the embeddings of all nodes into a single embedding vector, producing a representation from the states of all nodes to perform different tasks, such as graph classification and regression.

(3) \textbf{Local Model Uploading}: After local training, each client $i$ uploads its updated local model $\omega_{t,i}$ to the PS in round $t$, which allows the PS to collect the locally updated models from clients for aggregating a new global model.

(4) \textbf{Global Model Aggregation}: In this step, the PS aggregates the received local models to update the global model for the next round. The most commonly used aggregation method is weighted averaging \cite{mcmahan2017communication}, as below:

\begin{equation} \label{Agger_func}
 	\omega_{t+1} = \sum_{i=1}^{N}\frac{n_{i}}{n} \omega_{t,i}
\end{equation}

\subsection{Motivation for Framework Design}\label{subsec:motivation}
In FGL, graph data is characterized by two key aspects: structural properties and node features, both of which significantly influence model training performance. Structural properties offer a macroscopic perspective of the graph, capturing global properties and providing the model with insights into the overall structure. This facilitates the extraction of graph-invariant information, which is particularly valuable for understanding the overarching graph structure. However, these properties may overlook the node feature and the direct connectivity between neighboring nodes. In contrast, node features capture embeddings that represent the relationships between neighboring nodes and highlight the local features of individual nodes. This provides a more nuanced view of the node and its neighbors, complementing the broader perspective offered by structural properties. We further validate the impact of these two characteristics through experimental evaluation.

\begin{figure}[t]
    \centering
    \setlength{\abovecaptionskip}{-0.2mm}
    \subfigure[ChemBio Graph]{
        \includegraphics[width=1.8in]{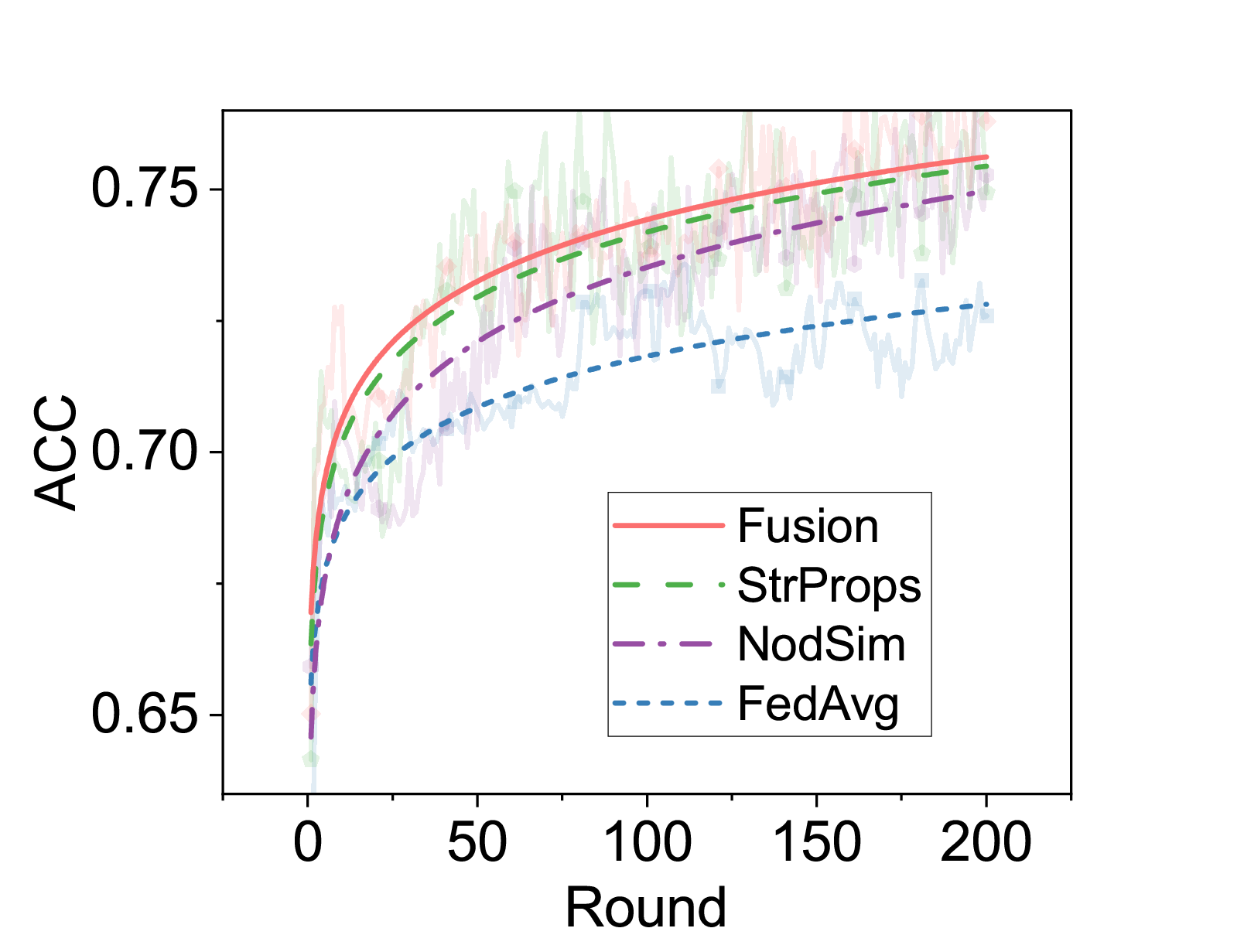}\label{subfig:test_chem_independent}
    }\hspace{-0.9cm}
    \subfigure[Social Networks]{
        \includegraphics[width=1.8in]{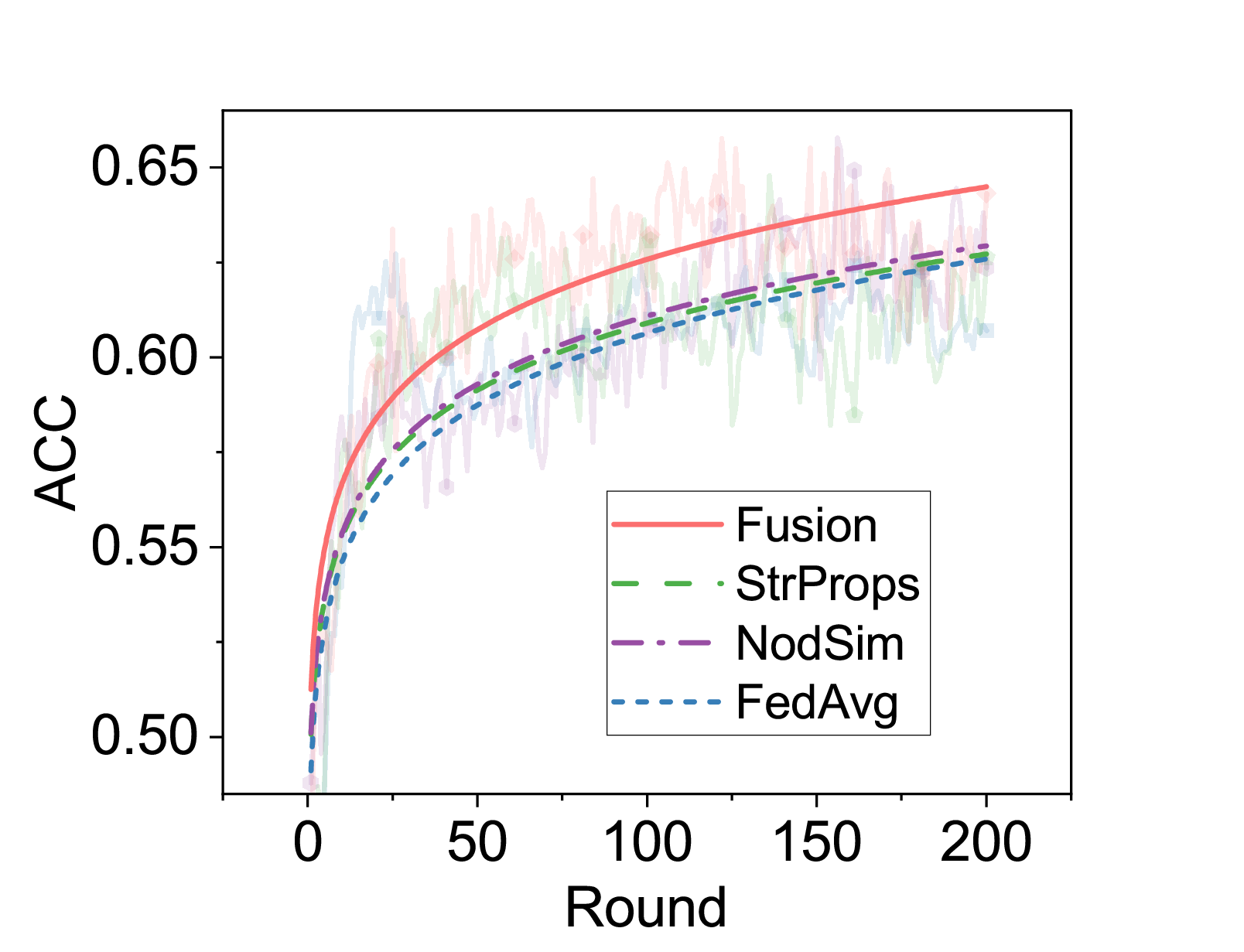}\label{subfig:test_sn_independent}
    }
    \caption{The impact of different data characteristics on two datasets.}
    \label{fig:test_independent}
    \vspace{-5mm}
\end{figure}

\textbf{(1) The fusion of structural properties and node features can contribute to training performance on various graph datasets.} 
We conduct a set of experiments to separately extract structural properties and node features, and then fuse them. 
We adopt two types of graph datasets, \ie, ChemBio Graph and Social Networks \cite{tan2023federated}. 
Each dataset contains multiple distinct samples, resulting in an inherent non-IID distribution due to variations in class distribution.
To leverage structural properties, we combine random walk results with the maximum degree of each node, incorporating these into the model input. 
This approach preserves privacy while extracting structural properties into the client's structural model. 
Collaborative learning among clients with similar graph structures enhances knowledge sharing and model training efficiency. Therefore, we divide the clients into three clusters based on their structural models.
Structural properties are then shared in each cluster to enhance the learning of graph structures.
For node features, we calculate model similarities to identify the shortest paths between clients, generating a common node model by aggregating local models from clients along the longest one in these paths. The final global model combines this common node model with the global model from all clients. Additionally, we evaluate the effect of fusing structural properties and node features in a 1:1 ratio.

For convenience, we use StrProps to indicate the approach utilizing only structural properties, NodSim for the approach using only node features, and Fusion for the approach fusing both types of characteristics. 
We use FedAvg \cite{mcmahan2017communication} as the baseline, a classical algorithm in FL, which updates the global model by weighted averaging the locally trained model parameters from clients. 
Due to significant fluctuations in the results, we apply a fitting function to smooth the data for easier comparison. Solid lines represent the fitted results, while semi-transparent lines denote the original experimental results.
As shown in Fig. \ref{fig:test_independent}, both StrProps and NodSim achieve better test accuracy than FedAvg across different types of datasets. For instance, on the ChemBio Graph dataset, StrProps has an accuracy improvement of 2.6\% compared to FedAvg after 200 rounds of model training. 

We also observe that Fusion leads to further improvement in the model test accuracy. On the Social Networks dataset, Fusion achieves a 1.6\% improvement in the model test accuracy after 200 training rounds compared to NodSim. Moreover, the performance gains vary between different datasets. On the ChemBio Graph dataset, StrProps demonstrates a more substantial performance improvement, whereas NodSim leads on the Social Networks dataset. This discrepancy arises because the ChemBio Graph dataset requires the classification of the entire graph, emphasizing structural properties. In contrast, the Social Networks dataset focuses on making predictions based on node relationships, underscoring node features.
This raises an important question: since both structural properties and node features impact performance differently across datasets, should we not adjust the combination ratio of these two characteristics to optimize their effectiveness in each case?

\begin{table}[t]
	\centering
	\caption{Model test performance for the impact of different graph characteristics combination ratios on two datasets.}\label{overall_performance}
	\begin{tabular}{c|c|c}
		\hline
		\multirow{2}{*}{StrProps:NodSim} & \multicolumn{1}{c|}{ChemBio Graph} & \multicolumn{1}{c}{Social Networks} \\ \cline{2-3}
		& avg. acc (\%) & avg. acc (\%) \\ \hline
		9:1 & 76.75 & 63.58 \\ 
		7:3 & \underline{\textbf{78.16}} & 64.26 \\ 
		5:5 & 76.50 & 64.32 \\ 
		3:7 & 76.15 & \underline{\textbf{65.23}} \\ 
		1:9 & 76.00 & 63.89 \\ \hline
	\end{tabular} \label{TABLE_test}
\end{table}

\textbf{(2) Different combination ratios need to be chosen for various graph datasets to achieve the optimal performance.} 
To further explore how different combination ratios between structural properties and node features impact model training, we conduct additional experiments to determine the extent to which their fusion enhances performance. 
In this set of experiments, we vary the combination ratio of structural properties and node features used, allowing us to evaluate their combined effectiveness across different datasets. 
We represent these ratios as varying degrees of fusion, for example, StrProps:NodSim=3:7 indicates a fusion of structural properties and node features at a ratio of 3:7.
By Table \ref{TABLE_test}, varying the combination ratios of structural properties and node features leads to the different model test accuracy. For instance, after 200 training rounds on the ChemBio Graph dataset, a 7:3 ratio of structural properties to node features improves model test accuracy by approximately 2.2\% compared to a ratio of 1:9. 
Moreover, the optimal combination ratio varies across different types of graph data. For example, the optimal combination ratio for the ChemBio Graph dataset is 7:3, while for the Social Networks dataset, it is 3:7. 
These results inspire us to leverage the structural properties and node features of graph data for comprehensive graph data learning while fusing them to improve the model training performance. Additionally, the variation in optimal combination ratios across different datasets suggests that each dataset requires a tailored combination ratio to achieve the best accuracy.

\section{Framework Design}\label{sec:framework}
\begin{figure*}[t]
    \centering
    \includegraphics[width=1\textwidth]{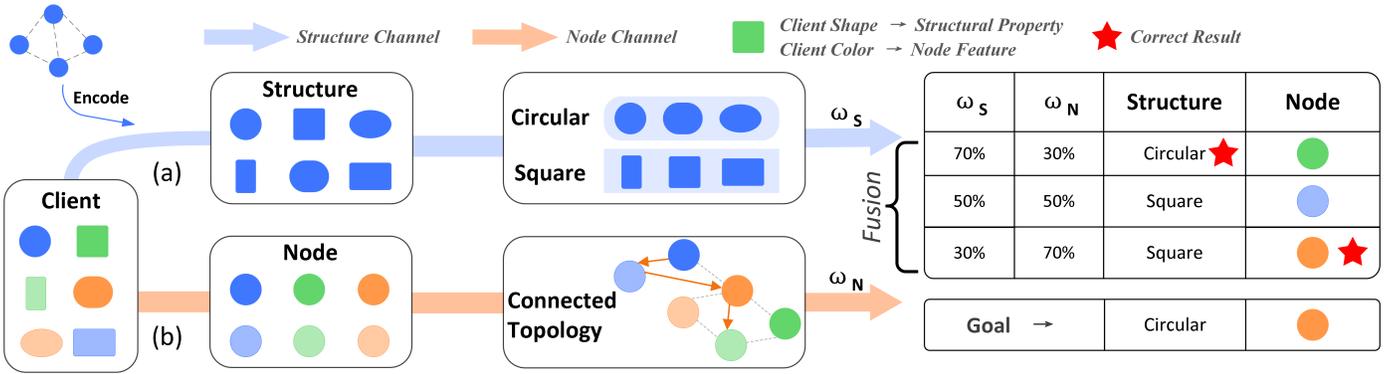}
    \caption{Illustration of the Graph Characteristics Extraction and Fusion step in FedGCF. (a) FedGCF executes the structural properties extraction process, where clients are clustered based on structural properties (\ie, client shape). The clustering results will be used to aggregate the shared structural model $\omega_S$ for different clusters. (b) FedGCF executes the node features extraction process, where the clients with the common node features are selected (clients with arrows passing through) based on node features (\ie, client color) to aggregate the common node model $\omega_N$. After both characteristics are extracted, they are fused based on the combination ratio to perform the prediction task.}
    \label{Agger_eee}
\end{figure*}

In this section, we introduce the FedGCF framework, which is designed to extract and utilize both the structural properties and node features of graph data. Similar to the workflow of FGL, FedGCF primarily involves four steps in each round of the global process: Local Model Training, Local Model Uploading, Graph Characteristics Extraction and Fusion and Model Distribution. In each round, each client initializes its local model based on the received model in the last round and proceeds with local training. After completing several iterations of local updates, the clients upload the updated models to the PS. The PS then extracts structural properties and node features from the uploaded models (Section \ref{subsec:Characteristics Extraction}). These extracted characteristics are subsequently fused on the PS using the proposed algorithm (Section \ref{subsec:MAB}) to produce the final characteristic fusion model. Once the characteristic fusion model is obtained, it is distributed to the respective clients. This process continues until the model converges.
The major differences between FGL and our proposed framework are evident in the processes of \textit{Local Model Training} and \textit{Graph Characteristics Extraction and Fusion}.

During local model training, in each round $t$, client $i$ preprocesses its local graph data to obtain basic graph structural information using the random walk algorithm and the maximum degree algorithm \cite{shun2013ligra,salihoglu2014optimizing,merrill2012scalable}. Both the raw graph data and the preprocessed graph data are utilized along two dimensions to train different GNN models, resulting in the trained local model $\omega_{t,i}$ and structural model $\omega_{s,i}$, respectively. The local model $\omega_{t,i}$, consistent with the traditional FGL local model, is used in the subsequent steps to extract node features from graph data. In contrast, the structural model $\omega_{s,i}$ is utilized to extract structural properties of graph data.

The Graph Characteristics Extraction and Fusion is composed of three key components, \ie, structural properties extraction, node features extraction and graph characteristics fusion. 
The foremost two components focus on characteristic extraction, while the graph characteristics fusion component fuses these characteristics with a proper ratio. 
Different from the traditional FGL workflow, we modify the role of the PS, which originally focuses solely on model aggregation and transmission tasks, and provide a detailed explanation of how the PS processes the models received from the clients to extract and fuse both characteristics.

\textbf{(1) Structural properties extraction:} In the structural properties extraction process, FedGCF focuses on the extraction of the unique structural properties inherent in graph data. 
Traditional GNN models rely on a message-passing mechanism to iteratively aggregate features from neighboring nodes \cite{gilmer2017neural,zhou2020graph}. This approach limits their ability to capture structural properties, as it focuses on a restricted subset of neighboring nodes, potentially overlooking broader graph structures.
Therefore, FedGCF extracts the structural properties from the structural model trained on graph structural information. Upon receiving the structural models, FedGCF clusters clients into different clusters based on their structural models. Within each cluster, the structural models from different clients are aggregated to produce a shared structural model that incorporates the extracted structural properties. The shared structural model obtained in each cluster will share the extracted structural properties within the cluster. This clustering and sharing mechanism enables clients with similar graph structures to mutually enhance their learning of structural properties, fostering a collaborative learning environment where the shared knowledge benefits all clients in the cluster.

\textbf{(2) Node features extraction:} 
In the node features extraction process, identifying common features from graph data is essential for building a generalized and effective model. 
We propose calculating model similarities between clients and mapping them as distances to form a connected topology. Here, each vertex represents a client, and the distances reflect node feature similarity, with longer distances indicating lower similarity.
To identify the most common node features, FedGCF determines the shortest paths between clients based on their distances. The vertices along these paths represent clients with similar data characteristics. From these paths, the path with the maximum distance is selected. We aggregate the local models from the clients along the path with the maximum distance to obtain the common node model, which includes the extracted common node features. 
By sharing these common node features among all clients, FedGCF significantly enhances the model’s generalization ability, ensuring strong performance across diverse graph datasets. This approach improves the model's generalization with data from different clients, facilitating cross-client collaboration and enabling more effective information sharing.

\textbf{(3) Graph characteristics fusion:} After both graph characteristics are extracted, FedGCF enters the graph characteristics fusion process. The combination ratio of structural properties and node features affects the understanding of graph data and the model training performance. FedGCF employs an effective algorithm (Section \ref{subsec:MAB}) that adaptively determines the current characteristics combination ratio. After determining the combination ratio, FedGCF combines the previously obtained shared structural models with the common node model to generate characteristic fusion models. Since the shared structural models are constructed independently for each client cluster, their number matches the number of client clusters, while the common node model remains a single entity across the entire process. Consequently, the resulting characteristic fusion models are formed by respectively combining each cluster’s shared structural model with the common node model, yielding a number of fusion models equal to the number of client clusters. Once the characteristic fusion models are obtained, they are distributed to the corresponding clients based on the client cluster they belong to, ensuring efficient utilization of structural properties and node features. This whole process will be repeated until the global model converges.

In order to illustrate the Graph Characteristics Extraction and Fusion step in FedGCF, we take Fig. \ref{Agger_eee} as an example. 
Assume there are six clients participating in training where shapes represent the structural properties of the graph data on the clients (gradually transitioning from circular to square), and colors represent the node features of the graph data on the clients.
In process (a), structural properties are extracted, clustering clients into circular or square categories based on their structural properties. Similar clients are clustered together, and the shared structural model $\omega_S$, aggregated in each cluster, is shared within the cluster. Process (b) focuses on node features extraction and calculates the model similarity between any two clients. A connected topology is constructed based on these similarities, and each vertex in this topology represents a client. The clients with the common node features selected based on the connected topology (clients with arrows passing through) are recorded. These clients are then utilized to aggregate their models into the common node model $\omega_N$.
Once both two characteristics have been extracted, they are fused on the PS with the combination ratio. 
For example, if structural properties and node features are fused at a 70\% to 30\% ratio, the structure prediction can accurately classify the shape as circular, however, the node prediction may fail to classify the color as orange.
Conversely, if the combination ratio is 30\% to 70\%, favoring node features, the node can correctly classify the color as orange, but may misclassify the structure as square.
As illustrated above, FedGCF can set a proper combination ratio for various characteristics based on the graph data characteristic emphasis, thereby enhancing the overall accuracy of the prediction results. 
However, as previously described, an improper combination ratio may lead to biases in understanding graph data as well as reduce result accuracy, and determining an appropriate combination ratio across different datasets remains an issue. 

\section{Algorithm Design}\label{sec:algorithm}

\subsection{Key Goals}\label{subsec:problem}
There are two key goals in implementing an adaptive graph characteristics fusion algorithm.

\textbf{Goal 1.} \textit{To fully leverage graph data, it is essential to simultaneously extract its structural properties and node features.} 
Due to the data composition patterns of graph data, there are inherent structural properties and node features. 
However, the existing FGL frameworks often struggle to fully extract and utilize these two types of characteristics. 
Firstly, these frameworks typically rely on the traditional GNN model, which encodes node representations based on features and aggregates information through the message-passing mechanism among neighboring nodes, thereby capturing only a limited portion of the structural information. 
These frameworks prioritize direct neighborhood interactions, limiting the representation of long-range dependencies and complex topological structures, which presents challenges in capturing unique structural properties in graph data. 
Secondly, most FGL methods do not fully utilize structural properties and node features, often overlooking the critical information in graph data. 
Directly aggregating GNN models trained on different data distributions cannot capture valuable graph characteristics across graph data, leading to a decrease in model training efficiency and test accuracy \cite{hsieh2020non,bojchevski2018netgan,reddi2020adaptive}. 
Consequently, the inadequate exploitation of structural properties and node features can result in significant information loss, thereby degrading model training performance and increasing the likelihood of errors in downstream tasks.

\textbf{Goal 2.} \textit{Effectively fusing the extracted structural properties and node features with appropriate combination ratios is critical for handling diverse graph scenarios.}
Different graph scenarios may emphasize different types of characteristics, making it crucial to effectively fuse the structural properties and node features. A mismatch between the combination ratio and the characteristic emphasis can lead to biased results or even incorrect conclusions. 
Although directly accessing graph data to understand the current data characteristics emphasis is the most straightforward approach, it is not permissible due to privacy constraints in FGL. 
Additionally, characteristic emphasis derived from historical experience or manually obtained is difficult to ensure complete accuracy. 
To address this challenge, parameters such as training outcomes can be used to infer the relative importance of structural properties and node features. However, this method provides only approximate guidance and lacks sufficient precision. 
Therefore, enhancing the effectiveness of these parameters is essential to achieve a more accurate combination ratio, ensuring that the fusion process better aligns with the data characteristics emphasis.

\begin{algorithm}[t]
    \caption{Process of PCE at round $t$} \label{alg_PCE}
    \textbf{Input:} The structural model $\omega_{s,i}$ and local model $\omega_{t,i}$ of each client $i$. The number of client clusters $K$ and the number of selected paths $P$. \\
    \textbf{Output:} The shared structural models $\omega_{S,k}, \forall k \in K$, the common node model $\omega_N$ and the client cluster set $C_K$.
    \begin{algorithmic}[1]
        \State Cluster clients into $K$ clusters through the structural model $\omega_{s,i}$ of each client $i$. {\color[RGB]{148,0,211}\algorithmiccomment{Structural properties extraction}} \label{alg1_1}
        \State Update the client cluster set $C_K$ according to the clustering results. \label{alg1_2}
        \For{each client cluster set $c_k \in C_K$} \label{alg1_3}
            \State Aggregate structural models $\omega_{s,i}, \forall i\in c_k$ to generate the shared structural model $\omega_{S,k}$ as Eq. (\ref{Agger_func}). \label{alg1_4}
        \EndFor \label{alg1_5}
        \State Calculate distance $d_{i,j}$ between clients $i$ and $j$ as Eq. (\ref{similarity_to_distance}). {\color[RGB]{148,0,211}\algorithmiccomment{Node features extraction}} \label{alg1_6}
        \State Build the connected topology for clients. \label{alg1_7}
        \State Calculate the shortest path between any two clients. \label{alg1_8}
        \State Select the top $P$ longest paths from these paths. \label{alg1_9}
        \State Record the clients involved in the selected paths as $P_C$. \label{alg1_10}
        \State Aggregate local models $\omega_{t,i}, \forall i\in P_C$ to generate the common node model $\omega_N$ as Eq. (\ref{Agger_func}). \label{alg1_11}
    \end{algorithmic}
\end{algorithm}

\subsection{Characteristics Extraction Algorithm}\label{subsec:Characteristics Extraction}
To achieve the first goal, we propose the Parallel Characteristic Extraction (PCE) algorithm, which simultaneously extracts and separates structural properties and node features from the graph data. The detailed steps are outlined in Alg. \ref{alg_PCE}. Specifically, PCE divides the complete characteristics extraction into two steps: structural properties extraction and node features extraction.

\textbf{Structural properties extraction.} To extract structural properties, PCE first preprocesses the raw graph data by adding a structural property vector for each node to bring out the graph structure. For each node $u$, in addition to its original node feature $x_u$, PCE generates a structural property vector $s_u$ using the graph traversal algorithms \cite{shun2013ligra,salihoglu2014optimizing,merrill2012scalable}, where $s_u$ encodes the structural information of node $u$. To obtain $s_u$, PCE utilizes both the random walk algorithm and the maximum degree algorithm \cite{xia2019random,tong2006fast,albalahi2022vertex}. The random walk algorithm captures the node's position and relationships within the graph, while the maximum degree algorithm reflects the node's connections. 
By merging the node feature $x_u$ and the structural property vector $s_u$, the node representation of each node becomes $\{x_u,s_u\}$. 
This additional dimension enriches the node's representation, enhancing its comprehensiveness and the quality of information captured for downstream tasks.


Although the structural property vector encapsulates the structural information of the graph data, directly using this vector as the structural properties can pose a risk of privacy leakage. 
To address this concern, PCE incorporates the structural property vector into the model training process while maintaining privacy. Specifically, PCE employs a GNN model to learn from the structural property vector and only utilizes the structural model $\omega_{s,i}$ obtained from each client $i$ to extract structural properties. 
Specifically, after receiving the structural model $\omega_{s,i}$ from each client $i$, the PS clusters these clients into $K$ clusters based on these models (Lines \ref{alg1_1}-\ref{alg1_2} in Alg. \ref{alg_PCE}). 
Since these models capture the structural property vectors, which reflect the graph data's structure on each client, clustering these clients based on their structural models effectively classifies the graph data across clients according to their graph structures.
PCE then independently processes each cluster $c_k \in C_K$. For each client $i \in c_k$, PCE aggregates the structural model $\omega_{s,i}$ to obtain the shared structural model $\omega_{S,k}$ as Eq. (\ref{Agger_func}). The resulting $K$ shared structural models contain the final available structural properties (Lines \ref{alg1_3}-\ref{alg1_5} in Alg. \ref{alg_PCE}).

\textbf{Node features extraction.} In this step, PCE processes the feature information of each client and constructs a connected topology, where the vertex represents the client and the distance between vertices reflects the similarity between client data features.
To quantify this similarity, PCE uses model similarity, denoted as $\sigma_{i,j}$, to represent the similarity of data features between clients $i$ and $j$. 
However, directly using $\sigma_{i,j}$ to represent the distance between any two clients is not appropriate because the range of $\sigma_{i,j}$ is too narrow (\eg, when using cosine similarity, the range of $\sigma_{i,j}$ is $\sigma_{i,j} \in [-1,1]$). 
Therefore, PCE adjusts $\sigma_{i,j}$ to represent the distance $d_{i,j}$ between clients $i$ and $j$ using the following formula:
\begin{equation} \label{similarity_to_distance}
	d_{i,j} = e ^ {\alpha\times(\frac{1-\sigma_{i,j}}{1+\sigma_{i,j}})} - 1
\end{equation}
where $\alpha$ is a constant value. The larger $\alpha$ is, the more sensitive the distance between clients becomes to the change in $\sigma_{i,j}$. The distance $d_{i,j}$ between clients $i$ and $j$ gradually increases as the model similarity between them decreases, and this distance exhibits an exponential increase as the similarity further decreases, amplifying the differences in features between various clients. Therefore, by mapping the similarity $\sigma_{i,j}$ to the distance between two clients using this function, the data feature similarity across clients can be effectively amplified (Lines \ref{alg1_6}-\ref{alg1_7} in Alg. \ref{alg_PCE}). After converting all clients into the connected topology, PCE can select clients with common data features using this connected topology. Specifically, PCE calculates the shortest path distance between any two clients and selects the top $P$ longest paths from these paths (Lines \ref{alg1_8}-\ref{alg1_9} in Alg. \ref{alg_PCE}). In the connected topology, vertex distances represent the similarity of node features between clients. Consequently, the shortest paths ensure that clients along each path share similar node features, while the top $P$ longest paths enable clients on these paths to cover the most common node features. After determining the clients with the most common node features, PCE aggregates the local model $\omega_{t,i}$ from each selected client $i$ to get the common node model $\omega_N$ as Eq. (\ref{Agger_func}) (Lines \ref{alg1_10}-\ref{alg1_11} in Alg. \ref{alg_PCE}). Then, by sharing the node features of these clients across all clients, the extraction and utilization of node features can be achieved.

\begin{algorithm}[t]
    \caption{Process of GCF at round $t$} \label{alg_AC}
    \textbf{Input:} The shared structural models $\omega_{S,k}, \forall k \in K$, the common node model $\omega_N$, the number of combination ratios $M$, the client cluster set $C_K$.\\
    \textbf{Output:} The characteristic fusion model.
    \begin{algorithmic}[1]
        \State Initialize the combination ratio score $\widehat{S}_m$ and the combination ratio selection counter $\widehat{N}_m^t$ for each combination ratio $m$ in the first round. \label{alg2_1}
        \State Update the cumulative reward $\widehat{R}_m$ for the selected combination ratio $m$ in the last round as Eq. (\ref{RE_m}). \label{alg2_2}
        \If{round $t \leq M+1$} \label{alg2_3}
            \State Randomly select a combination ratio and ensure each ratio is selected at least once in $M+1$ rounds. \label{alg2_4}
        \Else \label{alg2_5}
            \State Calculate the combination ratio score $\widehat{S}_m$ for each combination ratio $m$ as Eq. (\ref{SC_m}). \label{alg2_6}
            \State Select the combination ratio with the highest score. \label{alg2_7}
        \EndIf \label{alg2_8}
        \For{each client cluster set $c_k \in C_K$} \label{alg2_9}
            \For{each client $i \in c_k$} \label{alg2_10}
                \State Combine the shared structural model $\omega_{S,k}$ and the common node model $\omega_N$ with the selected combination ratio as the characteristic fusion model for client $i$. \label{alg2_11}
            \EndFor \label{alg2_12}
        \EndFor \label{alg2_13}
    \end{algorithmic}
\end{algorithm}

\subsection{MAB-Based Characteristics Fusion Algorithm}\label{subsec:MAB}
Due to the complexity of factors influencing the optimal combination ratio in FedGCF, such as data characteristics and model performance, it is challenging to determine the appropriate ratio prior to training. To address this, we introduce a Multi-Armed Bandit (MAB)-based algorithm, termed Graph Characteristics Fusion (GCF), which dynamically determines the optimal combination ratio in each round.
GCF estimates the expected reward for each combination ratio with minimal trials, enabling efficient exploitation of the optimal combination ratio. In each training round, the algorithm selects the combination ratio that maximizes the expected reward, ensuring the most beneficial fusion of structural properties and node features for model training.

The MAB problem, a classic in Reinforcement Learning (RL) \cite{dann2017unifying,schulman2015trust,schulman2017proximal}, addresses the trade-off between exploration and exploitation to maximize cumulative rewards \cite{auer2002finite,zhou2015survey,li2010contextual}. 
Formally, the MAB problem is formally represented as a tuple $(\mathcal{A}, \mathcal{R})$, where $\mathcal{A}$ denotes a set of $k$ actions $a$, and $\mathcal{R}$ represents the reward probability distributions. Each action $a_i \in \mathcal{A}$ is associated with a reward probability distribution $\mathcal{R}(r_i \mid a_i)$, from which a reward $r_i$ is drawn upon performing action $a_i$. The expected reward for each action is given by $Q(a) = \mathbb{E}[r \mid a]$, and the optimal expected reward is $Q^* = \max_{a \in \mathcal{A}} Q(a)$.
In FedGCF, determining the optimal characteristics combination ratio can be framed as an MAB problem. Here, each combination ratio represents an action, and the corresponding reward reflects this combination ratio's contribution to improving model training.

We show the GCF algorithm in Alg. \ref{alg_AC}. GCF first conducts several rounds of exploration to set the initial rewards for possible characteristics combination ratios (corresponding to the exploration in the MAB problem). 
Specifically, GCF presets $M$ groups of possible characteristics combination ratios. 
At the beginning of each round, GCF updates the existing variables based on the results of the previous round to ensure that the system can adapt in real time to the current environment (Lines \ref{alg2_1}-\ref{alg2_2} in Alg. \ref{alg_AC}). In the first round, GCF randomly selects a combination ratio from all ratios to fuse structural properties with node features. The random selection in the first round helps establish a baseline for determining the reward in subsequent explorations. After that, GCF runs $M$ rounds, each time randomly selecting a different characteristics combination ratio and using it to fuse both characteristics (Lines \ref{alg2_3}-\ref{alg2_4} in Alg. \ref{alg_AC}). The results after training with different characteristics combination ratios are then recorded using the following formula:
\begin{equation} \label{RE_m}
	\widehat{R}_m = 
        \begin{cases}
            \widehat{R}_m \times 0.9 + (e^{\beta \times \frac{r_t-r_b}{r_b}} - 0.99), & r_t \geq r_b\\
            \widehat{R}_m \times 0.9 - (e^{\beta \times \frac{r_b-r_t}{r_b}} - 0.99), & r_t < r_b\\
        \end{cases}
\end{equation}
where $\widehat{R}_m$ is the cumulative reward for the characteristics combination ratio $m$. Let $r_t$ represent the test accuracy in the current round $t$ and $r_b$ denote the best test accuracy in history. It is important to note that directly using the model's test accuracy after training as the reward for the selected combination ratio is not appropriate, as the test accuracy of the model tends to increase rapidly in the early rounds of training and then gradually stabilizes. This dynamic change conflicts with our need for stable rewards, making it difficult to directly use the test accuracy as a consistent reward. Therefore, we calculate the reward based on the ratio of the change between the test accuracy of each round and the best historical test accuracy, amplifying this ratio using an exponential function, where $\beta$ represents the amplification factor. The larger $\beta$ is, the more sensitive the reward is to the change. Additionally, GCF uses a cumulative method to update the reward after each selection. As the number of training rounds increases, we need to focus more on the model's current state rather than the historical experience. Accordingly, when determining the reward after each selection, we reduce the history reward, gradually decreasing the influence of early selections to focus more on the current changes.

After determining the basic reward for each combination ratio, GCF can begin to select the appropriate characteristics combination ratio based on these rewards. First, GCF calculates the score corresponding to each ratio according to the following formula:
\begin{equation} \label{SC_m}
	\widehat{S}_m = \frac{\widehat{R}_m}{\widehat{N}_m^t}+\sqrt{2\frac{\log t}{\widehat{N}_m^t}}
\end{equation}
where $\widehat{S}_m$ is the score and $\widehat{N}_m^t$ is the selected time in total $t$ rounds for combination ratio $m$ (Line \ref{alg2_6} in Alg. \ref{alg_AC}). The reward $\widehat{R}_m$ is calculated by Eq. (\ref{RE_m}). $\widehat{S}_m$ consists of the average reward and the selection balance. The average reward reflects the typical benefit obtained from each possible combination ratio. It provides an estimate of how effective each combination ratio is in contributing to overall model training. While the selection balance ensures that no combination ratio is neglected for extended periods. When a particular combination ratio remains unselected for a long time, the selection balance value gradually increases, prompting the system to reassess the selection priority. This selection balance continues to rise until it meets the preset selection criteria, thereby preventing any combination ratio from remaining unselected for an extended period of time.

GCF selects the combination ratio with the highest score as the selected combination ratio (Line \ref{alg2_7} in Alg. \ref{alg_AC}). Then, for each client $i \in c_k$, GCF calculates its characteristic fusion model with the shared structural model $\omega_{S,k}$ and the common node model $\omega_N$ based on the selected combination ratio (Lines \ref{alg2_9}-\ref{alg2_13} in Alg. \ref{alg_AC}). 
Finally, each client derives the local model based on the received characteristic fusion model in the next round. The local model obtained by each client continues training on the local graph dataset. The entire process repeats until the global model converges.

\section{Experimental Evaluation}\label{sec:evaluation}
\subsection{Datasets and Models}\label{subsec:dataset_model}
\textbf{Datasets}: We use three sets of benchmark datasets, \ie, Small Molecules (BZR, COX2, DHFR, PTC$\_$MR, AIDS, NCI1), Social Networks (COLLAB, IMDB-BINARY, IMDB-MULTI) and  MIX (BZR, COX2, DHFR, DD), which are commonly adopted in FGL \cite{xie2021federated,tan2023federated}. The Small Molecules dataset emphasizes leveraging the overall structural properties of graph data, where each graph represents a chemical molecule. The primary task for model training on this dataset is to predict whether a molecule exhibits certain bioactivity or chemical properties. In contrast, the Social Networks dataset places greater emphasis on utilizing node features. This dataset consists of different individuals (\eg, scientists in COLLAB, actors in IMDB-BINARY and IMDB-MULTI), and the task involves leveraging the collaboration and interaction connections among individuals to determine the research domain (COLLAB) or movie type (IMDB-BINARY and IMDB-MULTI). The MIX dataset is composed of a selection from the Small Molecules dataset (BZR, COX2, DHFR) and a Bioinformatics dataset (DD). These datasets are commonly combined for testing model classification accuracy in graph classification tasks \cite{xie2021federated,tan2023federated}. All four datasets come from the field of bioinformatics, supporting drug development and disease treatment, and involve the analysis of graph structures. Additionally, the MIX dataset, a combination of datasets from two different domains, can be used to evaluate the ability of various algorithms to train the effective GNN model across different domain graph data.

\begin{table*}[t]
	\centering
	\caption{Performance of FedGCF and baselines on the three datasets. We present the average test accuracy and the average gain over Local Training for FedGCF and baselines across five replicate experiments.}\label{overall performance}
	\centering
	\begin{tabular}{c|cc|cc|cc}
		\hline
		\multirow{2}{*}{Method} & \multicolumn{2}{c|}{Small Molecules} &\multicolumn{2}{c|}{Social Networks} & \multicolumn{2}{c}{MIX}\\\cline{2-7}
		& avg.acc (\%) & avg.gain (\%) & avg.acc (\%) & avg.gain (\%) & avg.acc (\%) & avg.gain (\%)\\\hline
        Local Training &73.55 &--- &61.97 &--- &69.92 &---\\\hline
		FedAvg &71.44 &-2.11 &60.13 &-1.84 &68.54 &-1.38\\\
		FedProx &71.97 &-1.58 &60.40 &-1.57 &68.38 &-1.54\\\
		GCFL &72.14 &-1.41 &58.52 &-3.45 &68.57 &-1.35\\\
        FedStar &74.69 &1.14 &63.42 &1.45 &75.68 &5.76\\\hline
        FedGCF &\textbf{76.38} &\textbf{2.83} &\textbf{65.76} &\textbf{3.79} &\textbf{77.49} &\textbf{7.57}\\\hline
	\end{tabular}
        \label{TABLE_overall}
\end{table*}

\textbf{Models}: We use two distinct models to encode structural properties and node features respectively, \ie, a three-layer GCN model for node features and a three-layer GIN model for structural properties \cite{kipf2016semi,xu2018powerful}. 
The GCN model applies a linear transformation to node features and then averages these transformed features within each node's neighborhood. However, it is limited by its method of averaging neighbor features, which may fail to distinguish some graph structures. The GIN model aggregates neighbor features by summation, preserving more structural information. To construct the structural property vectors for the graph data, we combine two types of structural encoding: one is based on the node’s maximum degree to represent degree embeddings, and the other is based on position embeddings calculated from a random walk algorithm. The node features of the graph data do not require any additional processing and remain consistent with the traditional FGL framework.

\subsection{Evaluation Setup}\label{subsec:experiment_setup}

\textbf{Baselines:} We adopt five different baselines for performance comparison: (1) \emph{Local Training} lets each client train the GNN model locally without any communication or data sharing. (2) \emph{FedAvg} \cite{mcmahan2017communication} is the most classical FL algorithm by averaging model parameters across clients and weighting them according to their graph data sizes to promote efficient model training. (3) \emph{FedProx} \cite{li2020federated} addresses the data heterogeneity in traditional FL by introducing a proximal term. This term prevents local models on clients from deviating too much relative to the global model. (4) \emph{GCFL} \cite{xie2021federated} reduces the graph data heterogeneity in FGL by leveraging node features to assign all the clients to multiple clusters with lower data heterogeneity through a clustering algorithm. (5) \emph{FedStar} \cite{tan2023federated} shares the underlying structural information in FGL by defining and encoding structural embeddings with the independent structural encoder. This structural information is then leveraged by sharing the encoder across all clients.

\textbf{Performance Metrics:} We mainly employ the following metrics to evaluate the performance of different algorithms: (1) Test Accuracy. In each round, we will evaluate the global model on the test dataset and record the test accuracy. (2) Communication Cost. We record the communication cost throughout the entire training process. The communication cost is divided into two parts: the effective payload of parameter uploads/downloads for each client during communication with PS in each round and the size of the distributed local models during deployment.

\textbf{Platform and Parameter Settings:} Our experiments are conducted on a deep learning platform. This workstation is equipped with an Intel(R) Xeon(R) Platinum 8358P, 4 NVIDIA RTX A6000 GPUs and 512 GB RAM.
We use the hidden feature size of 64 for all baselines. We use a batch
size of 128 and the Adam optimizer with a learning rate of 0.001 \cite{kingma2014adam,reddi2019convergence}. For the two types of structure encodings for the structural properties: a node degree embedding based on the maximum degree algorithm and a positional embedding based on the random walk algorithm, we set the dimensions of both to 16. To ensure convergence, we train 200 rounds for all baselines. We divide the dataset into a ratio of 8:1:1 for the training set, the validation set and the test set.


\subsection{Overall Performance Comparison}\label{subsec:Experimental setup}

We first present the accuracy performance of FedGCF and the baselines across the three datasets, as shown in Table \ref{TABLE_overall}. The results demonstrate that FedGCF outperforms all baselines.
On the Small Molecules and Social Networks datasets, FedGCF achieves significant improvements in average accuracy, with increases of 4.94\% and 7.24\% compared to FedAvg and GCFL, respectively. Additionally, FedGCF outperforms Local Training, with average accuracy gains of 2.83\% and 3.79\%. On the MIX dataset, client data is more heterogeneous due to the dataset composition from different domains. As a result, most baselines show significant performance degradation and fail to surpass Local Training. 
However, FedGCF still achieves an impressive average accuracy improvement of 7.57\%. 
It also outperforms FedStar, which shares structural properties, by 1.81\%. On the MIX dataset, most baselines perform poorly due to the high degree of data heterogeneity. 
This heterogeneity is reflected in the diverse characteristics of each graph and leads to difficulties in effectively utilizing the structural properties of graph data across clients.
However, FedGCF not only classifies data effectively by leveraging the graph’s structural properties within each dataset, but also enables the sharing of exploitable structural information among clients with similar data structures. Additionally, FedGCF extracts common local node features from different graph data and shares them across all clients. 
Subsequently, FedGCF leverages the shared structural properties and common node features, achieving outstanding accuracy performance.

\begin{table*}[t]
	\centering
	\caption{Performance of FedGCF and baselines on different data distributions.}
	\centering
	\begin{tabular}{c|ccc|ccc}
		\hline
		\multirow{2}{*}{Method} & \multicolumn{3}{c|}{Small Molecules} &\multicolumn{3}{c}{Social Networks}\\\cline{2-7}
		& IID(\%) & non-IID-0.5(\%) & non-IID-0.7(\%) & IID(\%) & non-IID-0.5(\%) & non-IID-0.7(\%)\\\hline
		FedAvg &71.44 &74.20 &75.05    &60.13 &60.31 &60.31\\\
		FedProx &71.97 &73.97 &75.40   &60.40 &60.21 &61.39\\\
		GCFL &72.14 &74.29 &75.97  &58.52 &59.36 &59.35\\\
        FedStar &74.69 &76.11 &79.01  &63.42 &62.46 &63.28 \\\hline
        FedGCF &\textbf{76.38} &\textbf{80.00} &\textbf{81.29} &\textbf{65.76} &\textbf{66.02} &\textbf{65.98}\\\hline
	\end{tabular}
        \label{TABLE_noniid}
\end{table*}

\begin{figure}[t]
    \centering
    \setlength{\abovecaptionskip}{-0.2mm}
    \subfigure[Test Accuracy]{
        \includegraphics[width=1.8in]{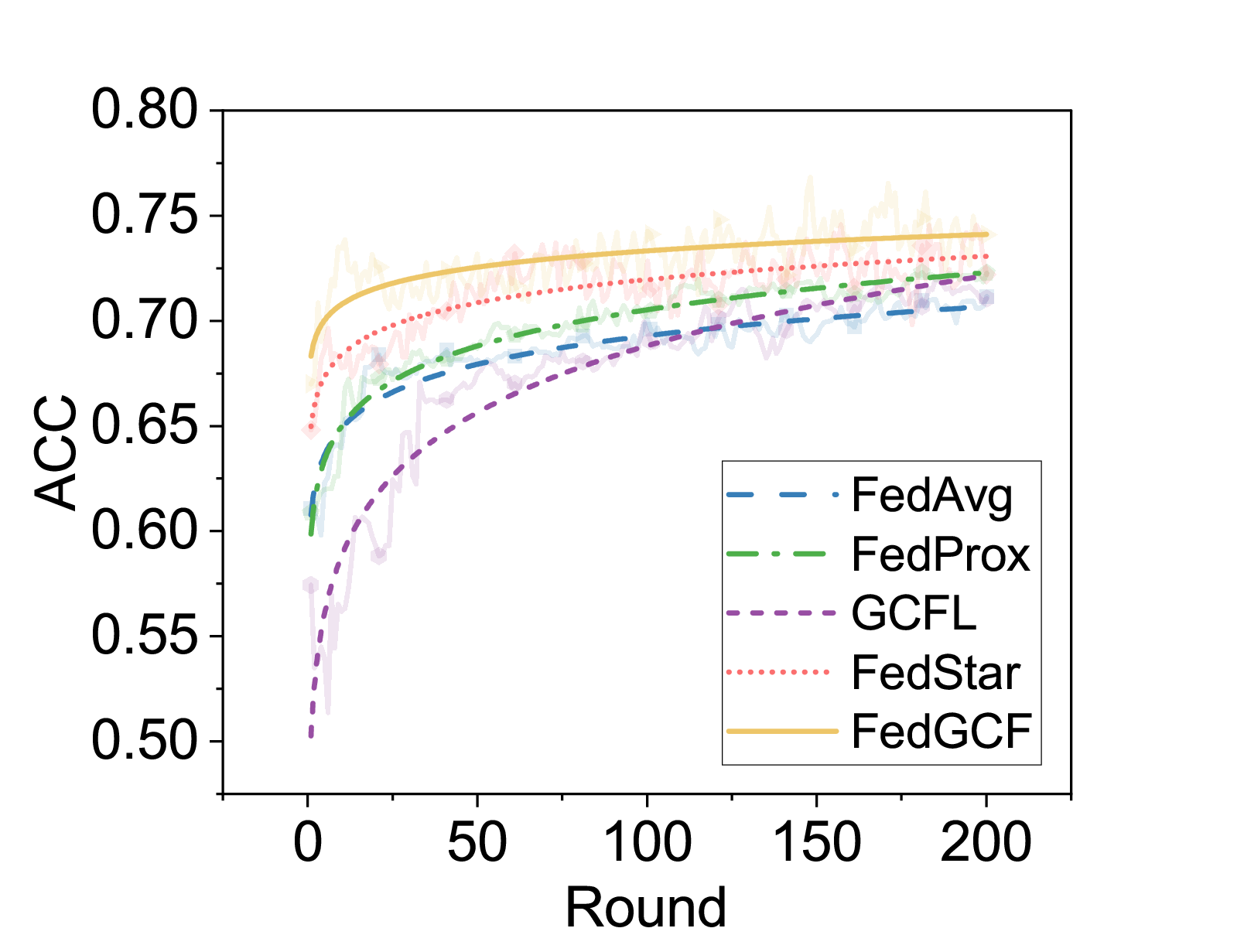}\label{subfig:init_chem_acc}
    }\hspace{-0.9cm}
    \subfigure[Communication Cost]{
        \includegraphics[width=1.8in]{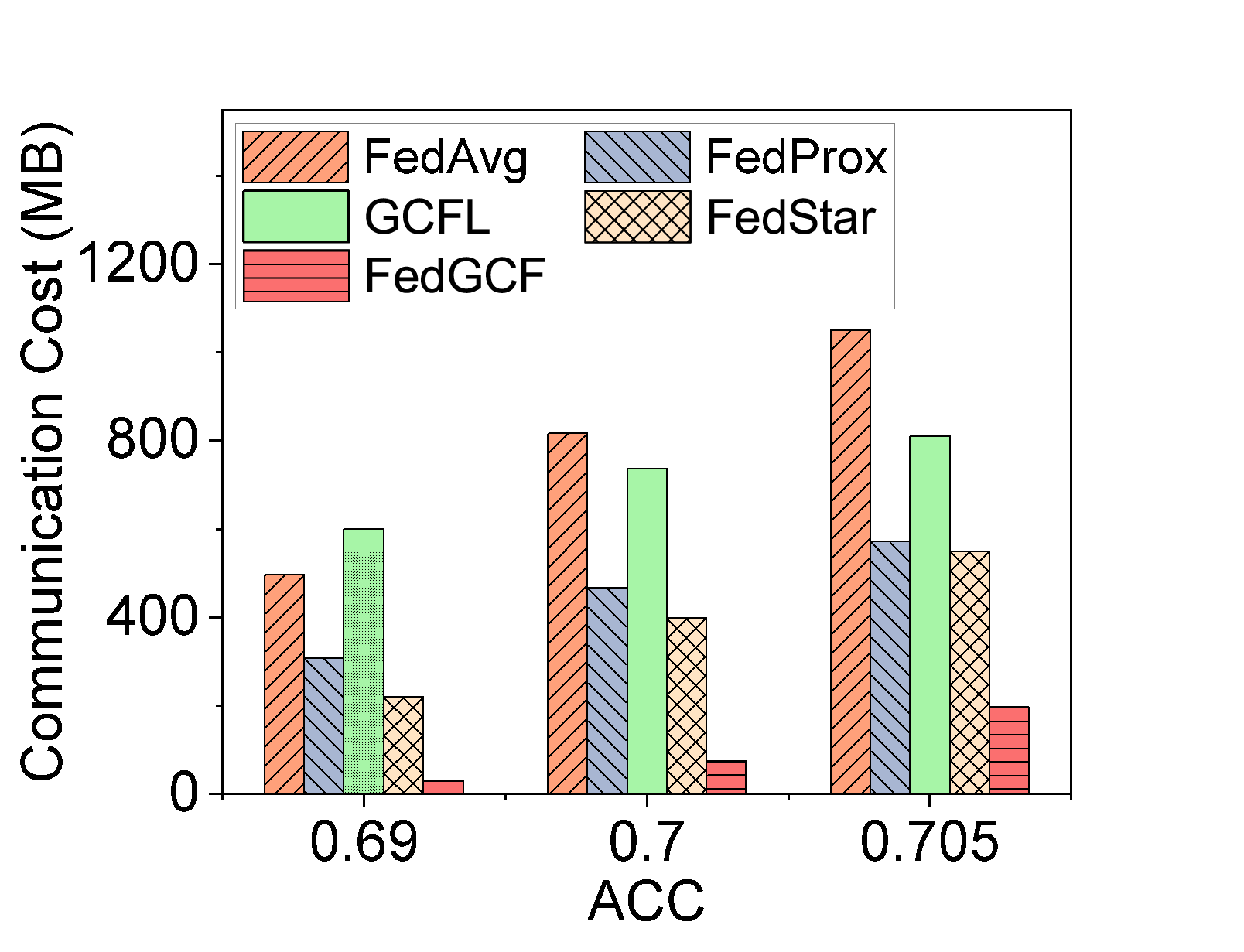}\label{subfig:init_chem_communication}
    }
    \caption{The test accuracy and communication cost for FedGCF and baselines on the Small Molecules dataset.}
    \label{fig:init_experiment_chem}
    \vspace{-5mm}
\end{figure}

\begin{figure}[t]
    \centering
    \setlength{\abovecaptionskip}{-0.2mm}
    \subfigure[Test Accuracy]{
        \includegraphics[width=1.8in]{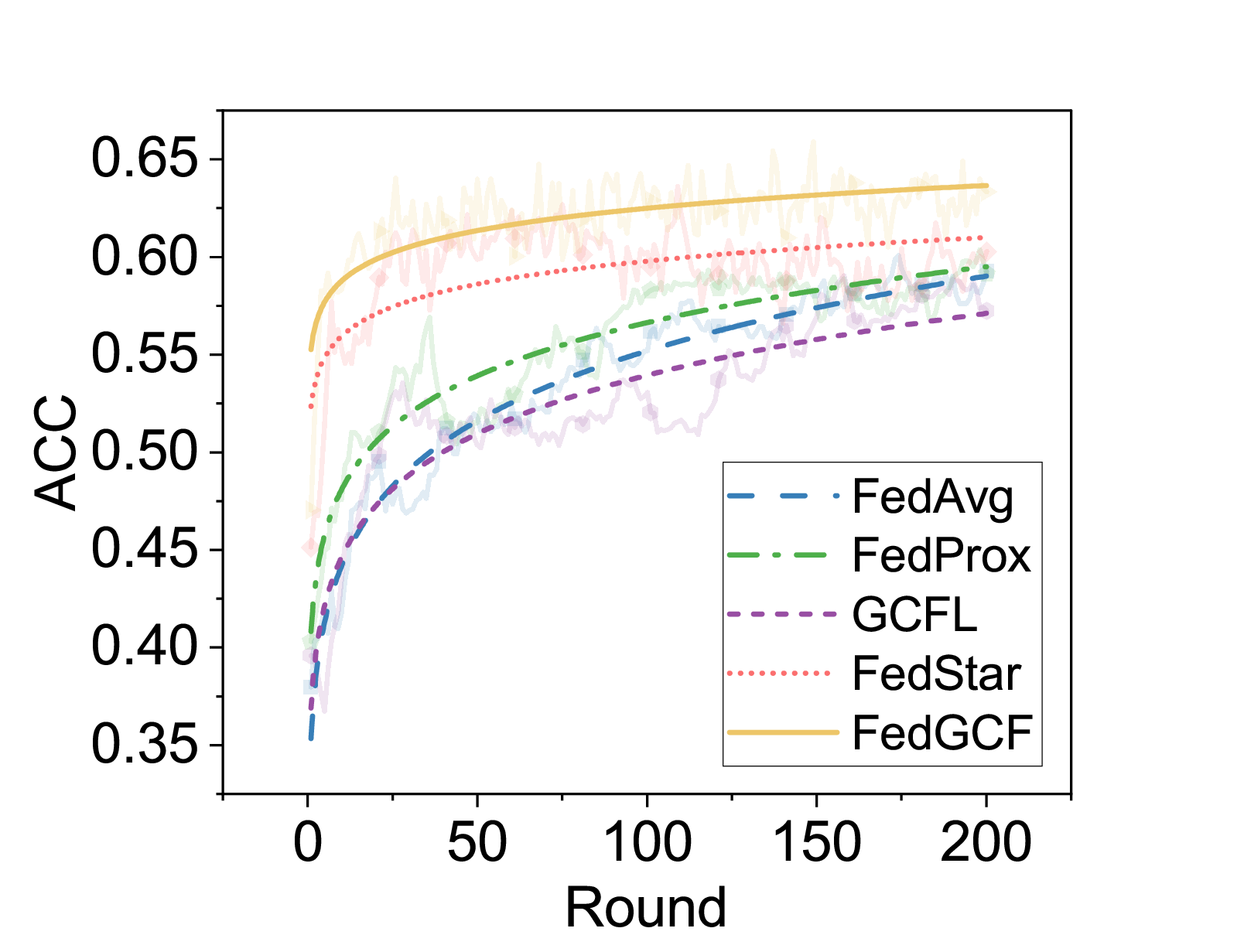}\label{subfig:init_sn_com}
    }\hspace{-0.9cm}
    \subfigure[Communication Cost]{
        \includegraphics[width=1.8in]{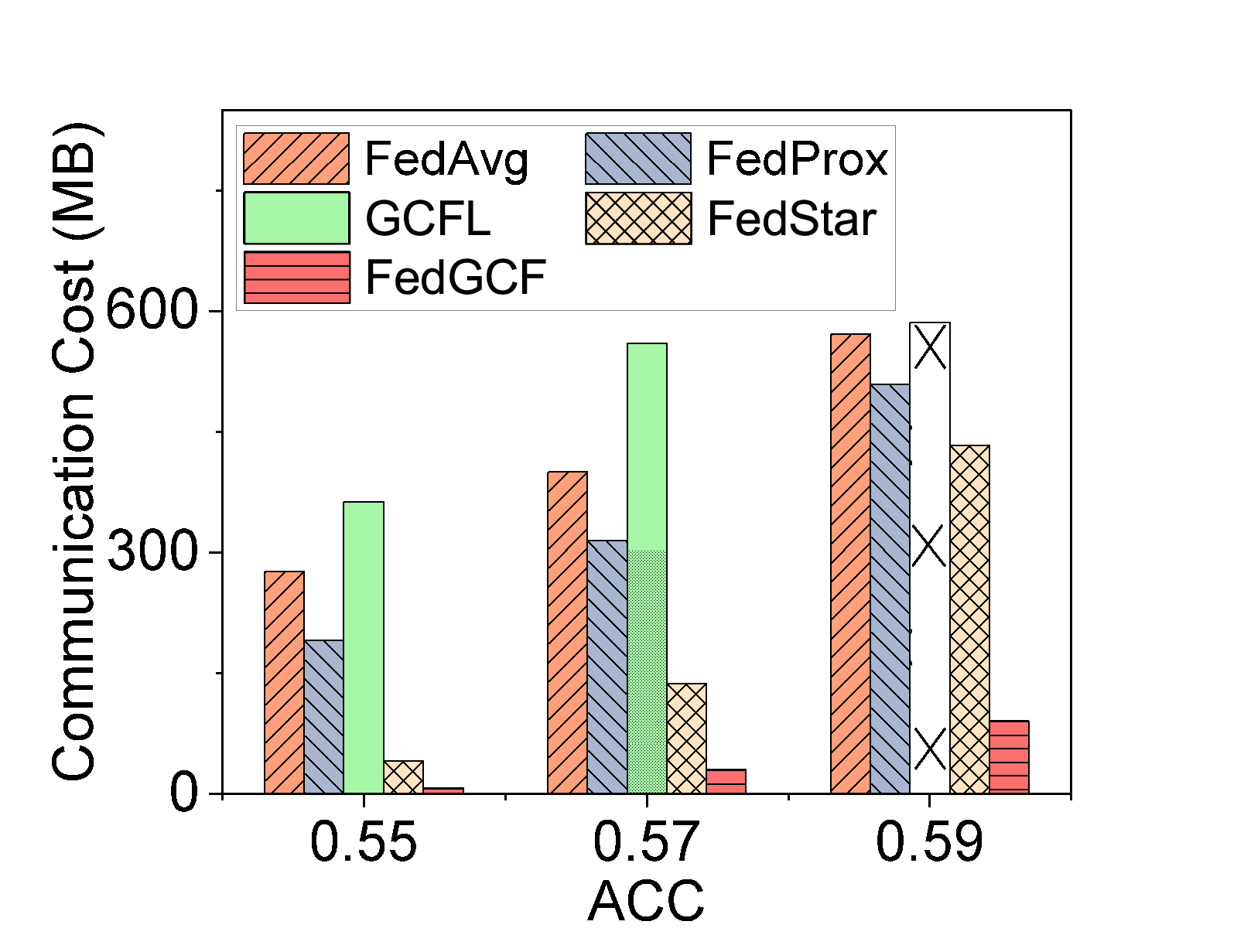}\label{subfig:init_sn_communication}
    }
    \caption{The test accuracy and communication cost for FedGCF and baselines on the Social Networks dataset (the baseline below the accuracy threshold is marked with the cross).}
    \label{fig:init_experiment_sn}
    \vspace{-5mm}
\end{figure}

\begin{figure}[t]
    \centering
    \setlength{\abovecaptionskip}{-0.2mm}
    \subfigure[Test Accuracy]{
        \includegraphics[width=1.8in]{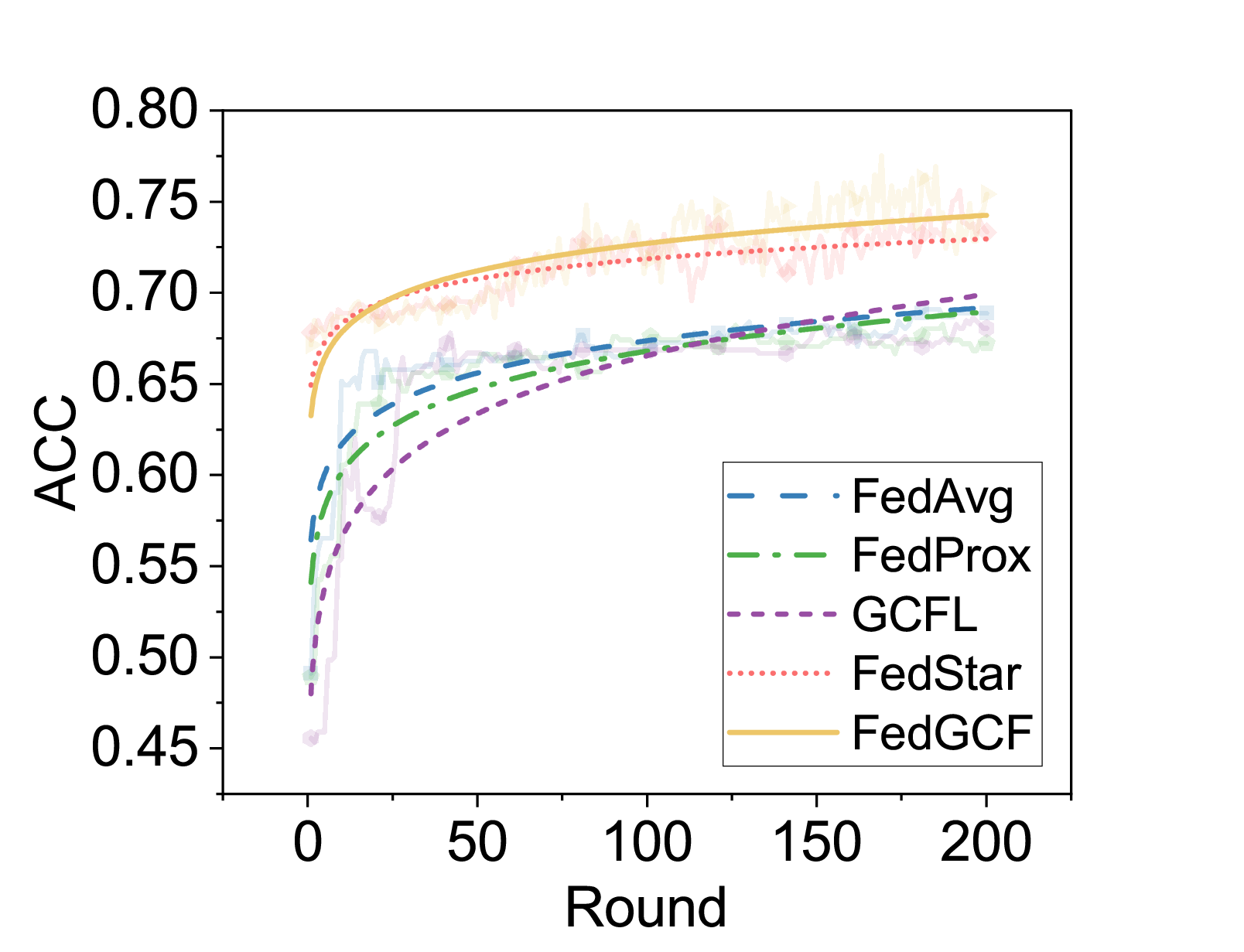}\label{subfig:init_mix_acc}
    }\hspace{-0.9cm}
    \subfigure[Communication Cost]{
        \includegraphics[width=1.8in]{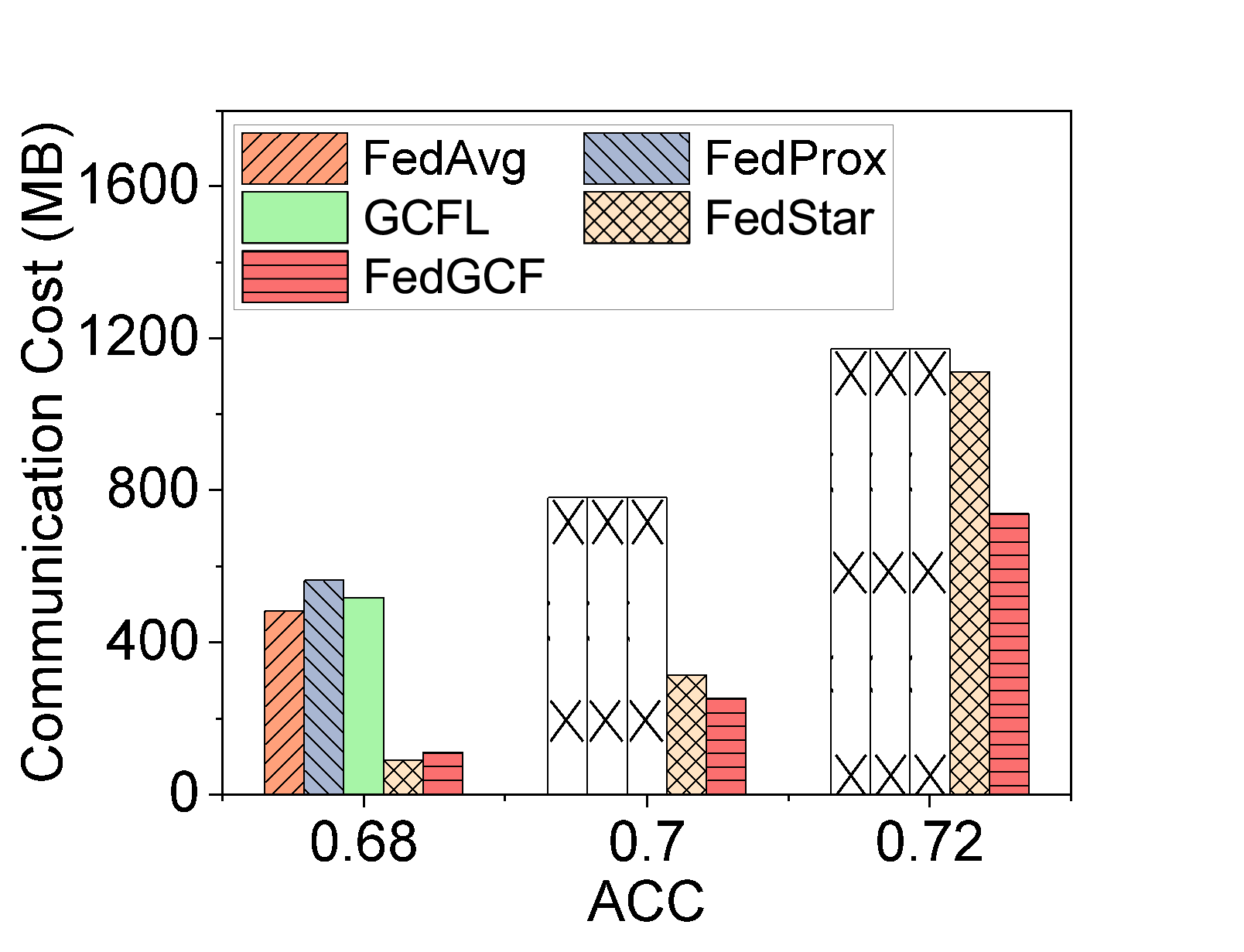}\label{subfig:init_mix_communication}
    }
    \caption{The test accuracy and communication cost for FedGCF and baselines on the Mix dataset (the baselines below the accuracy threshold are marked with the cross).}
    \label{fig:init_experiment_mix}
    \vspace{-5mm}
\end{figure}

In the left plots of Figs. \ref{fig:init_experiment_chem}-\ref{fig:init_experiment_mix}, we present the test results after training the model over varying numbers of rounds on different datasets. Due to significant fluctuations in the outcomes, we apply a consistent fitting function to smooth the test results. The solid lines represent the fitted results, while the semi-transparent lines depict the actual results. Overall, FedGCF consistently outperforms other baselines in test accuracy across all datasets and demonstrates a faster convergence rate. For example, on the Small Molecules dataset, FedGCF achieves a test accuracy of 72\% after 32 rounds, whereas FedStar and GCFL require 103 and 195 rounds, respectively. Similarly, on the MIX dataset, FedGCF reaches a 72\% test accuracy after 73 rounds, while FedStar needs 110 rounds, 50.7\% more than FedGCF. Additionally, none of the other baselines can achieve 72\% model test accuracy. These results highlight the efficiency and effectiveness of FedGCF in accelerating the model convergence and improving performance compared to the baselines. The substantial improvement highlights that our method effectively extracts and utilizes structural properties at each training round, thereby maximizing the benefits of characteristic sharing in FGL and accelerating convergence.

To further evaluate the resource efficiency of FedGCF, we record the communication costs of FedGCF and the baselines when achieving different model target test accuracy across various datasets, as shown in the right plots of Figs. \ref{fig:init_experiment_chem}-\ref{fig:init_experiment_mix}. Evidently, FedGCF can reach the target model test accuracy with the least communication cost compared with the baselines. Specifically, FedGCF uses only 18.75\% of the communication cost to reach a 70.5\% test accuracy on the Small Molecules dataset compared to FedAvg. For FedStar, FedGCF also reduces the communication cost by 64.18\%. On the MIX dataset, where client data exhibits extremely high heterogeneity, FedGCF requires only 111MB to achieve a 68\% test accuracy, while FedProx needs 564MB. This outstanding resource utilization efficiency stems from the efficient and adaptive fusion of graph data structural properties and node features in FedGCF. By globally sharing richly informative data characteristics, each client can fully learn the common characteristics within the dataset, thereby endowing FedGCF with high communication efficiency throughout the entire FGL training process.

\begin{figure}[t]
    \centering
    \setlength{\abovecaptionskip}{-0.2mm}
    \subfigure[Small Molecules]{
        \includegraphics[width=1.8in]{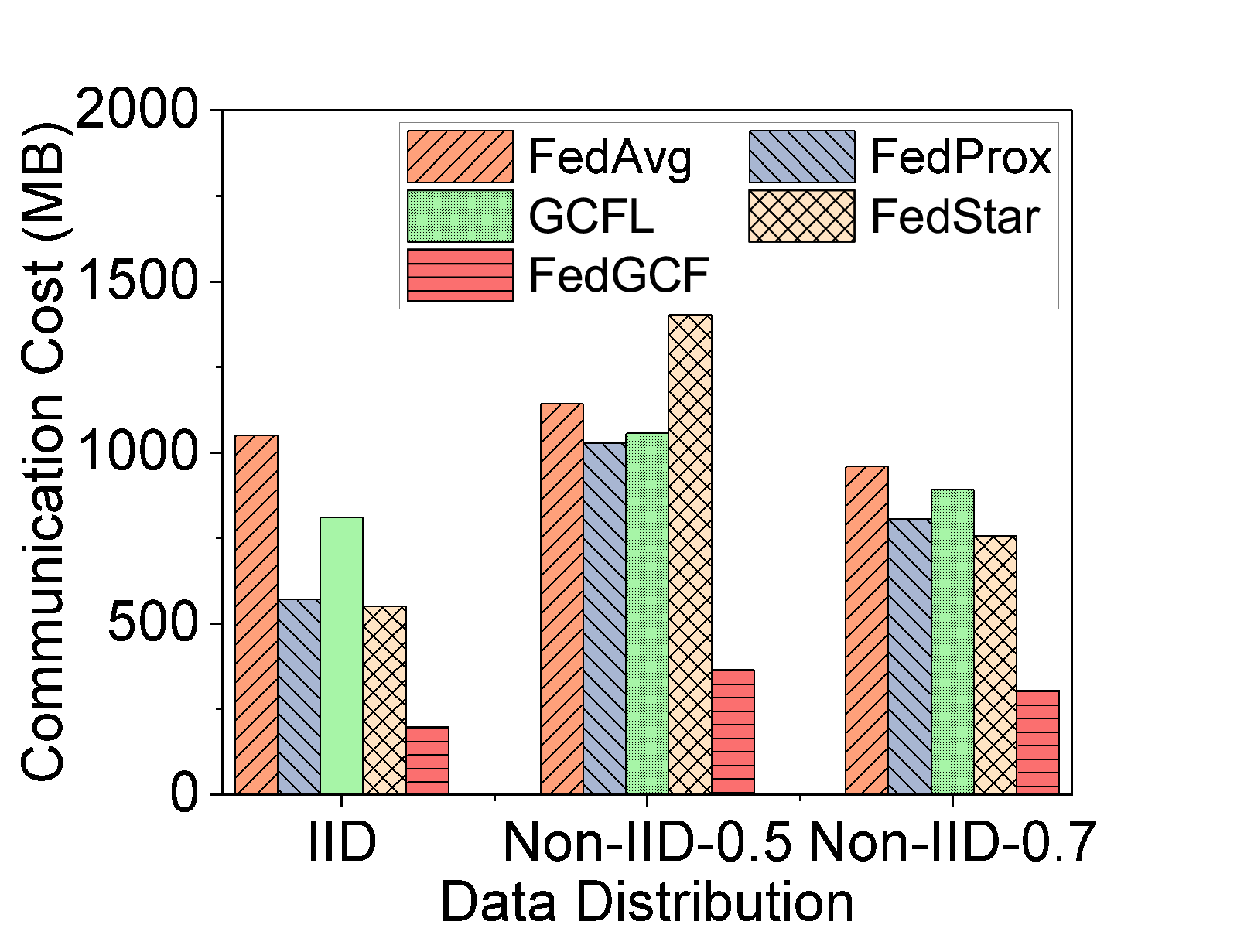}\label{subfig:noniid_sn_acc_50}
    }\hspace{-0.9cm}
    \subfigure[Social Networks]{
        \includegraphics[width=1.8in]{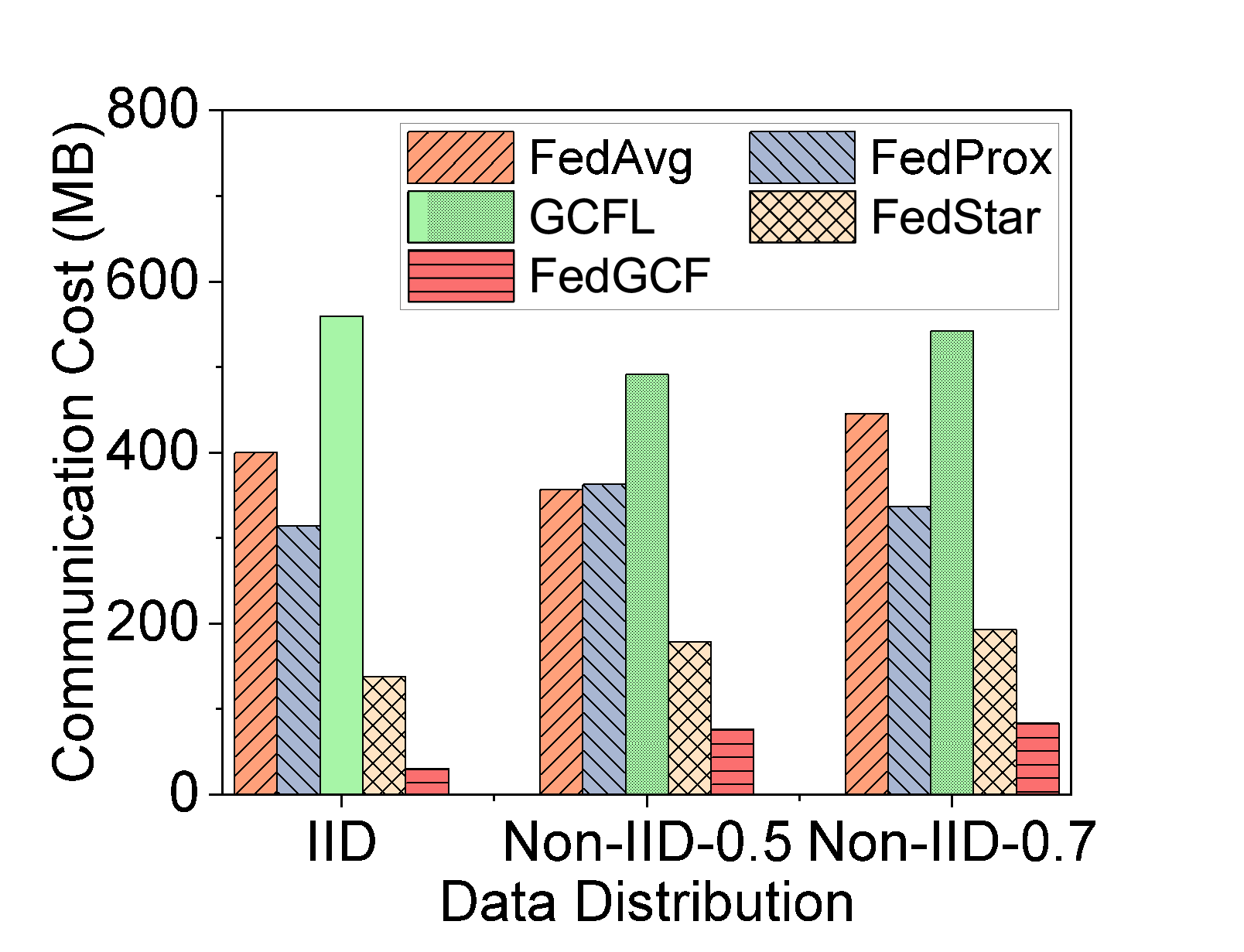}\label{subfig:noniid_sn_acc_70}
    }
    \caption{Communication cost to reach different target model test accuracy (Small Molecules: 70\%, 74\% and 75\%; Social Networks: 57\%, 58\% and 59\%) on the two datasets under different data distributions.}
    \label{fig:noniid_experiment_sn}
    \vspace{-5mm}
\end{figure}

\subsection{Adaptability to Data Distribution}\label{subsec:Data Distribution}
In this section, we conduct experiments using the Small Molecules and Social Networks datasets to evaluate the impact of data distributions on the test accuracy, with results shown in Table \ref{TABLE_noniid}. The models are trained on Small Molecules and Social Networks with three different data distributions, \ie, IID, non-IID-0.5 and non-IID-0.7. Specifically, under the IID setting, each client randomly selects graph data from the dataset to form its local graph data. Under the non-IID-0.5 setting, 50\% of the data on the clients consists of a single randomly selected class from the dataset, and the remaining 50\% is randomly selected. Similarly, under the non-IID-0.7 setting, 70\% of the data on the clients comes from a single randomly selected data class, while the rest client graph data is randomly selected.

In contrast to traditional FL, where test accuracy typically declines as data distribution becomes more skewed, models trained across both datasets here exhibit improved test accuracy with increasing data skew. 
This improvement stems from the model’s ability to learn a dominant data class more effectively. When clients randomly select data with 50\% or 70\% dominance of a particular class, more representative classes are likely to be included, leading to enhanced test performance as the proportion of a single class in each client's dataset grows.
In addition to these overall insights into different data distributions, we now focus on a direct comparison of test accuracy between FedGCF and the baseline methods. 
For the same data distributions, FedGCF consistently achieves the highest test accuracy, outperforming the four baselines by 4.94\% to 7.24\%. For instance, on the Small Molecules dataset, FedGCF achieves final test accuracy of 80.00\% and 81.29\% under the non-IID-0.5 and non-IID-0.7 data settings, respectively, whereas FedProx and FedAvg attain only 73.97\% and 75.05\% under different non-IID settings.

Furthermore, as data distribution becomes increasingly skewed, test accuracy variations on the Social Networks dataset are less pronounced than those on the Small Molecules dataset. Nevertheless, FedGCF continues to show some improvement over the IID setting. 
This indicates that in highly complex graph data, extreme data heterogeneity may hinder the model's ability to capture both structural and node characteristics. Meanwhile, baseline methods like FedProx and FedStar show performance declines under certain data distributions, highlighting their limitations in handling highly heterogeneous graph data distributions effectively.

In Fig. \ref{fig:noniid_experiment_sn}, we present the required communication costs to achieve the target model test accuracy across different datasets under three data distributions. 
Notably, FedGCF consistently achieves the target model accuracy with the lowest communication cost across all distributions compared with baselines. 
This efficiency is attributed to FedGCF's ability to identify and leverage shared characteristics within graph data while adaptively adjusting the combination ratio of structural properties and node features based on the characteristics of graph data.
Even as data skew increases, it can efficiently extract and share characteristics to ensure optimal model performance. 
For instance, on the Small Molecules dataset under the non-IID-0.7 data distribution, FedGCF requires only 303MB to achieve a 75\% test accuracy, whereas FedStar incurs 757MB.
Similarly, on the Social Networks dataset with the non-IID-0.5 data distribution, FedGCF reduces communication costs by 57.61\% and 79.10\% compared to FedStar and FedProx, respectively, to achieve a 58\% test accuracy. 
This reduction in communication costs highlights FedGCF’s ability to effectively manage resource consumption while achieving high accuracy, even in scenarios with significant data heterogeneity.

\begin{figure}[t]
    \centering
    \setlength{\abovecaptionskip}{-0.2mm}
    \subfigure[Social Networks]{
        \includegraphics[width=1.8in]{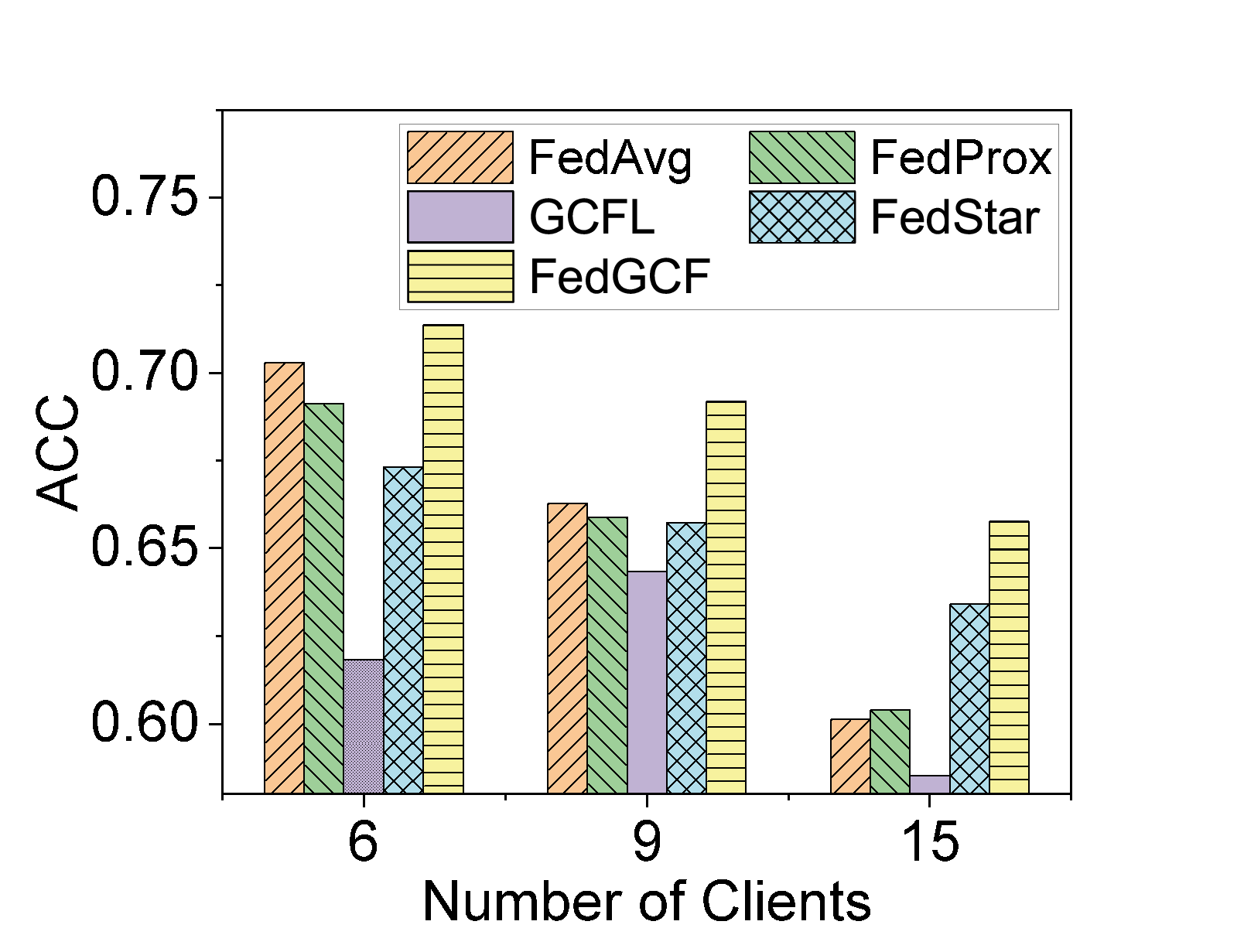}\label{subfig:client_number_communication_chem}
    }\hspace{-0.9cm}
    \subfigure[MIX]{
        \includegraphics[width=1.8in]{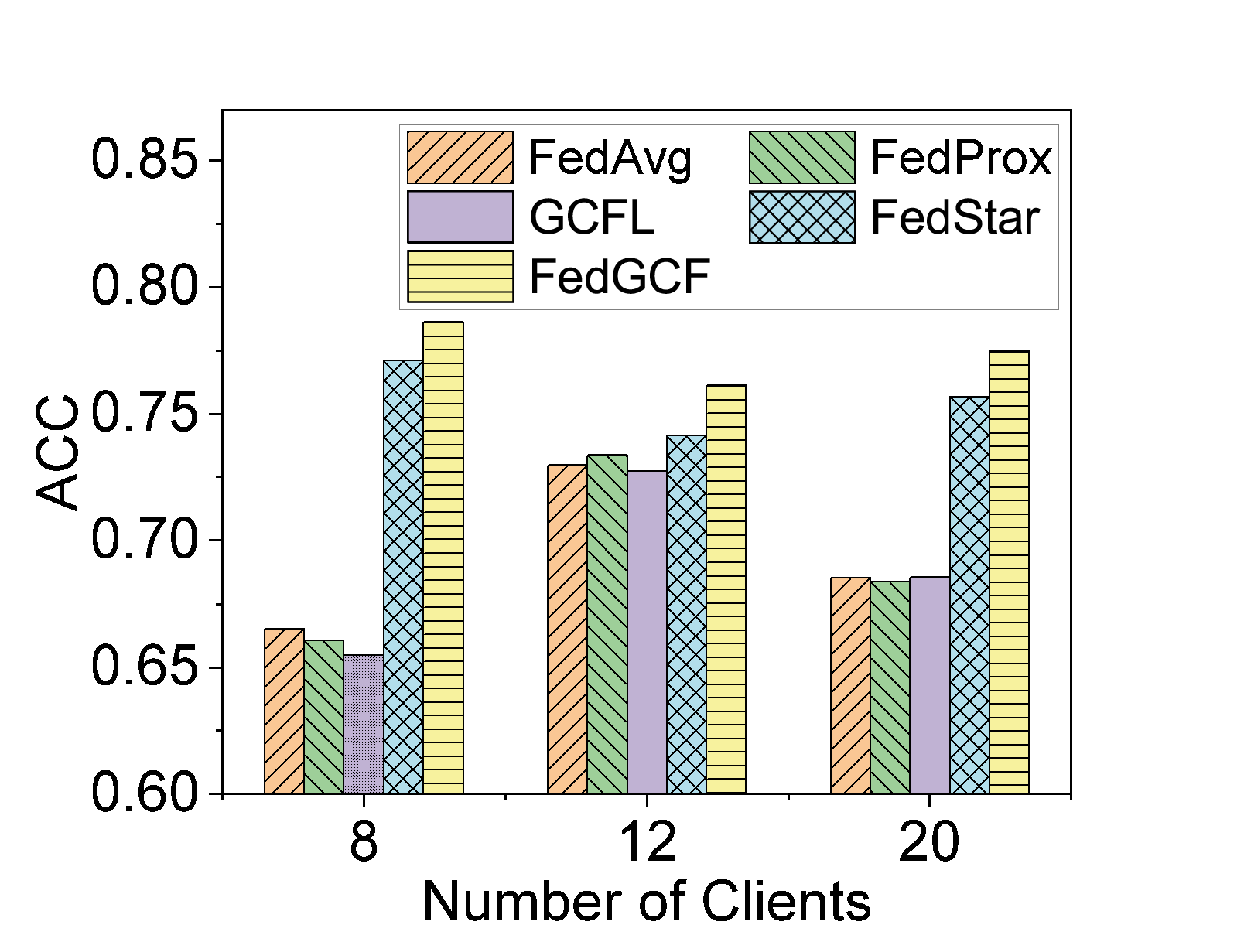}\label{subfig:client_number_communication_mix}
    }
    \caption{The test accuracy achieved by FedGCF and baselines on the two datasets under various numbers of clients.}
    \label{fig:clientnumber_experiment_acc}
    \vspace{-2mm}
\end{figure}

\subsection{Scalability with Client Number}\label{subsec:Client Number}
We evaluate the scalability of FedGCF with different numbers of clients. 
In this set of experiments, we vary the number of clients participating in training for FedGCF and baselines. 
We compare the test accuracy achieved by FedGCF and the baselines after 200 rounds of training on the Social Networks and MIX datasets, with the number of clients ranging from 6 to 20, as shown in Fig. \ref{fig:clientnumber_experiment_acc}.

Overall, as the number of participating clients decreases, the test accuracy achievable by the model under each baseline shows varying degrees of improvement, particularly on the Social Networks dataset. This effect arises because a smaller client count allows each client to access a richer dataset, thereby enhancing the model’s learning capability. 
Under varying client numbers, FedGCF consistently outperforms the baselines, achieving a test accuracy improvement of 1.53\% to 13.16\% over other methods.

Further analysis of the experimental results reveals that on the highly heterogeneous MIX dataset, methods like FedAvg and GCFL experience a significant performance drop when fewer clients participate. This outcome is counterintuitive, as a reduced number of clients would typically enhance or at least not decrease the performance of these methods. This decline, however, can be attributed to FedAvg and GCFL's limitations in effectively capturing and summarizing the graph characteristics across the dataset. Both methods struggle to generalize shared characteristics, leading to suboptimal representation of common graph information. Consequently, their inability to effectively extract and utilize the graph characteristics results in reduced overall performance.
In contrast, FedGCF captures both structural and node characteristics to share highly common information across clients effectively. Therefore, when applied to highly heterogeneous datasets like MIX, FedGCF maintains strong performance, achieving up to a 13.16\% improvement over GCFL on the MIX dataset.

\subsection{Component-wise Evaluation}\label{subsec:ablation experiment}
Finally, we comprehensively evaluate the effectiveness of the different components in FedGCF. Specifically, FedGCF is designed to separately extract structural properties and node features from graph data, and then adaptively fuse these two types of learned graph characteristics based on the current system state in the proper combination ratio. Thus, to thoroughly evaluate the impact of each component, we implement three distinct segmented versions of FedGCF, each isolating one of these key components. This approach allows us to assess and gain deeper insights into the individual effectiveness and contributions of these three key components to the overall model performance.

\begin{table}[t]
	\centering
	\caption{Performance of FedGCF and three segmented versions.}
	\centering
	\begin{tabular}{@{\hspace{0.1em}}c@{\hspace{0.1em}}|c@{\hspace{0.2em}}c@{\hspace{0.1em}}|c@{\hspace{0.2em}}c@{\hspace{0.1em}}}
		\hline
		\multirow{2}{*}{Method} & \multicolumn{2}{c|}{Small Molecules} &\multicolumn{2}{c}{Social Networks}\\\cline{2-5}
		& avg.acc(\%) & avg.gain(\%) & avg.acc(\%) & avg.gain(\%) \\\hline
		\hspace{0.05em}SC &74.90 &-1.48 &63.56 &-2.20\\\
        \hspace{-0.3em}NP &74.66 &-1.72 &64.55 &-1.21\\\
		\hspace{-0.3em}EF &75.64 &-0.74 &64.98 &-0.78\\\hline
        FedGCF &\textbf{76.38} &--- &\textbf{65.76} &---\\\hline
	\end{tabular}
        \label{TABLE_ablation}
\end{table}

\begin{figure}[t]
    \centering
    \setlength{\abovecaptionskip}{-0.2mm}
    \subfigure[Small Molecules]{
        \includegraphics[width=1.8in]{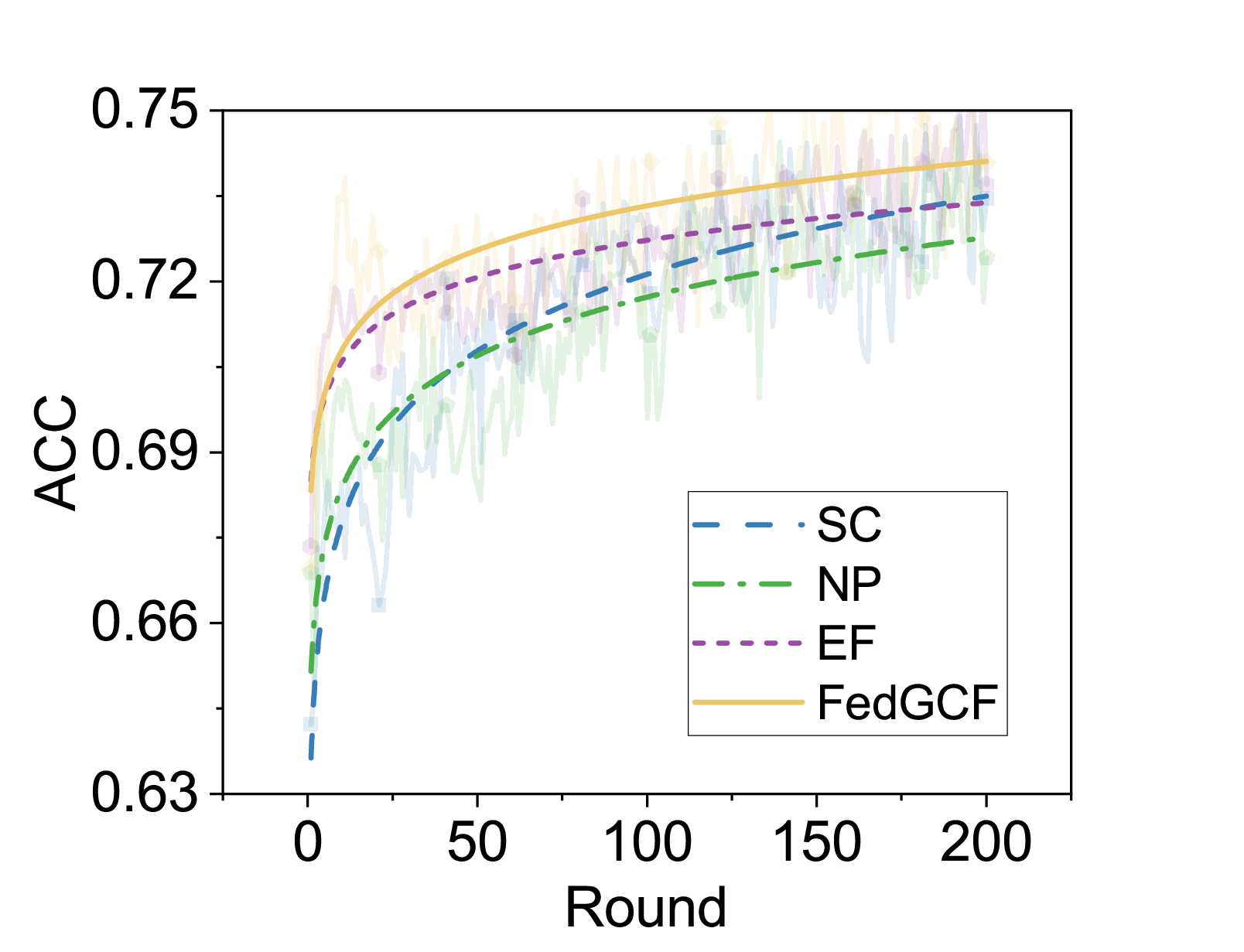}\label{subfig:ablation_acc_chem}
    }\hspace{-0.9cm}
    \subfigure[Social Networks]{
        \includegraphics[width=1.8in]{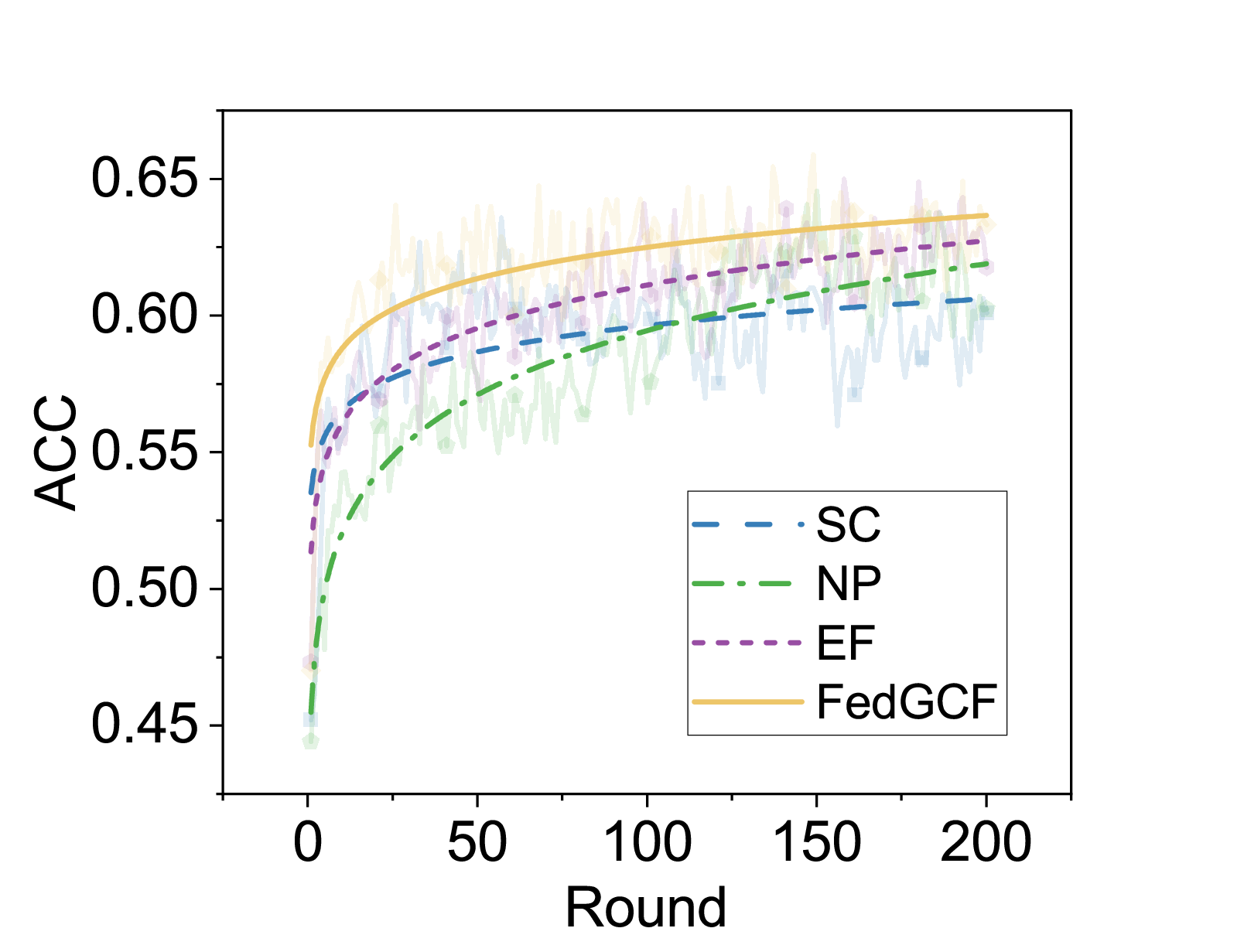}\label{subfig:ablation_acc_sn}
    }
    \caption{The model test accuracy of FedGCF and three segmented versions under different rounds on the various datasets.}
    \label{fig:ablation_acc}
    \vspace{-5mm}
\end{figure}

\begin{figure}[t]
    \centering
    \setlength{\abovecaptionskip}{-0.2mm}
    \subfigure[Small Molecules]{
        \includegraphics[width=1.8in]{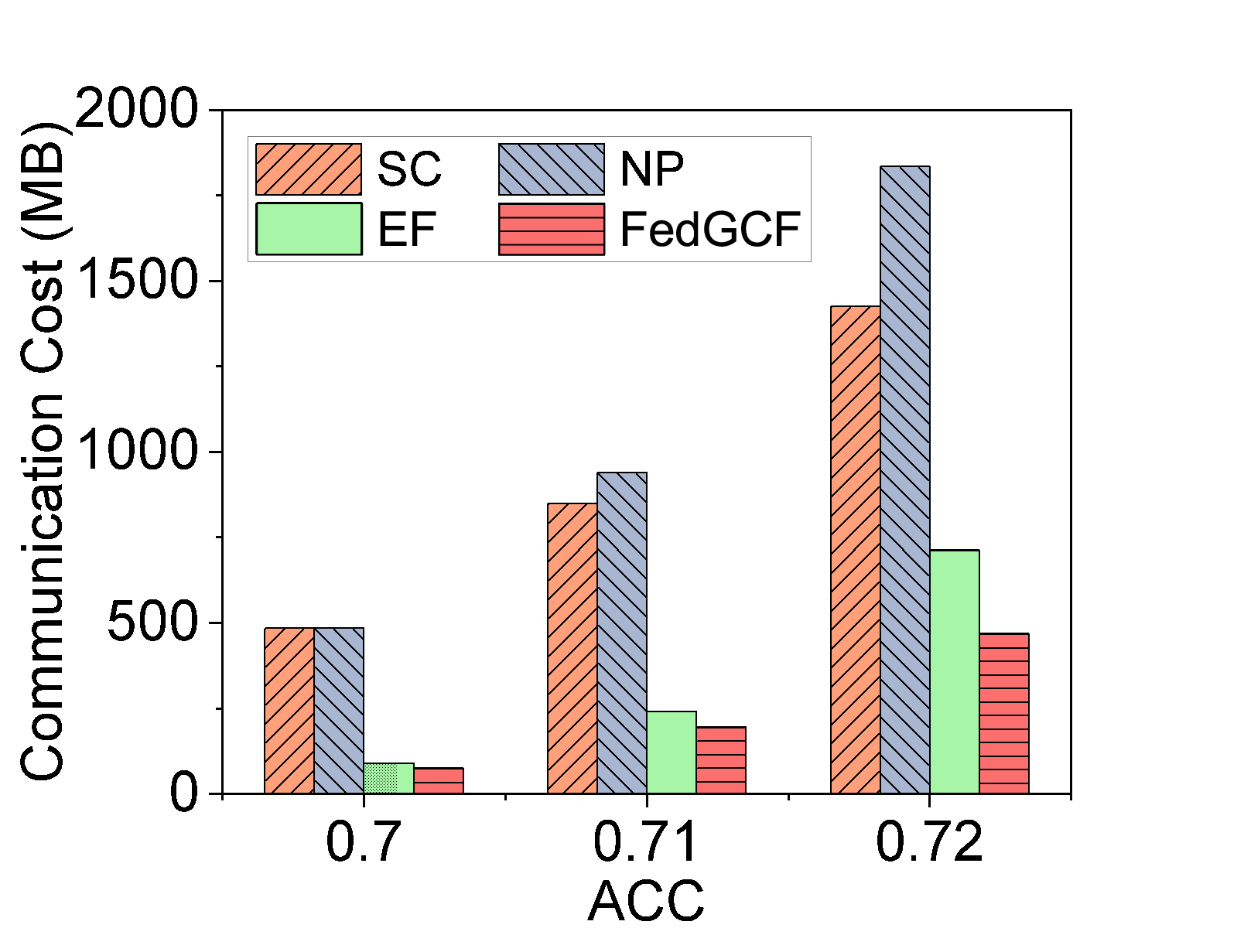}\label{subfig:ablation_communication_chem}
    }\hspace{-0.9cm}
    \subfigure[Social Networks]{
        \includegraphics[width=1.8in]{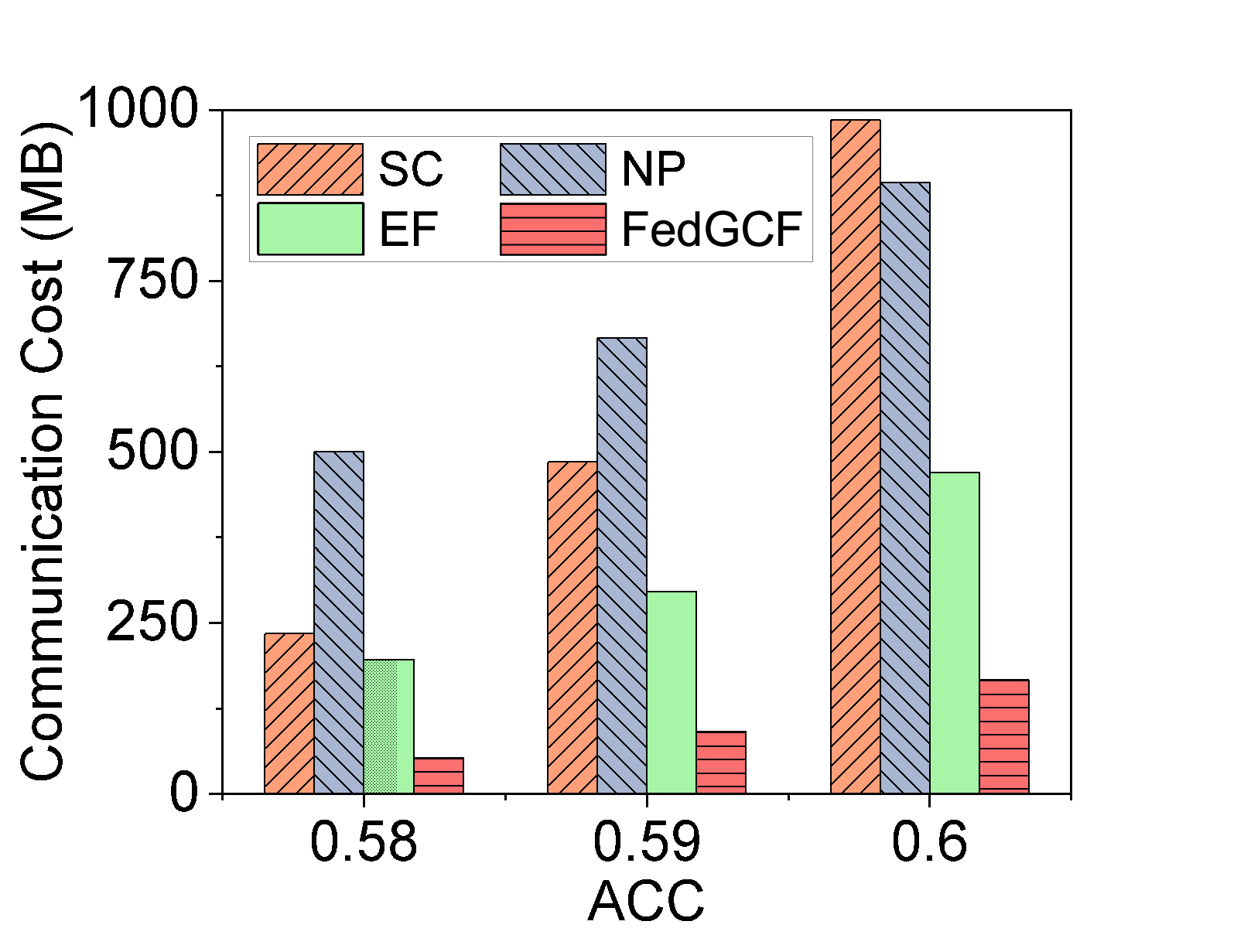}\label{subfig:ablation_communication_sn}
    }
    \caption{Communication cost to reach different target model test accuracy of FedGCF and three segmented versions.}
    \label{fig:ablation_communication}
    \vspace{-5mm}
\end{figure}

(1) \textbf{FedGCF w/o Node features extraction (SC)}. We disable the component for obtaining node features. In each round, the PS only clusters clients based on the obtained structural properties and shares similar structural properties among each cluster. As a result, FedGCF can only capture the overall structural properties of the graph data and knows very little about the specific connection characteristics between nodes, making it difficult to achieve an efficient model in more complex graph data.

(2) \textbf{FedGCF w/o Structural properties extraction (NP)}. We disable the component for obtaining structural properties. Specifically, the PS can only rely on the node features of the graph data to select the clients that cover the most common node features, sharing the node features among all the clients. Consequently, FedGCF lacks sufficient understanding of the structural information of the graph data, making it difficult to learn the overall structure of the graph data.

(3) \textbf{FedGCF w/o Graph characteristics fusion (EF)}. We disable the graph characteristics fusion part after obtaining structural properties and node features. Instead of adaptively adjusting the combination ratio based on the current system status, the PS fuses structural properties and node features equally at a 1:1 ratio. Therefore, the application of structural properties and node features in FedGCF will be rather casual, making it challenging to fully reflect the benefit of fusing these two types of characteristics.

FedGCF effectively extracts and utilizes structural properties and node features from graph data, adaptively fusing these two types of characteristics based on the current system environment to enhance the model test performance and communication resource efficiency. We first present the model test accuracy achieved by three different versions of FedGCF after training for 200 rounds on the Small Molecules and the Social Networks datasets, as shown in Table \ref{TABLE_ablation}. Intuitively, FedGCF clearly achieves the best results, while EF obtains the second-best performance. This is because, although EF utilizes both structural properties and node features, it combines these two types of characteristics rigidly with a fixed ratio, rather than adapting to the current system environment and training dynamics. Furthermore, NP and SC achieve the worst results on both datasets, however, NP performs worse than SC on the Small Molecules dataset but better on the Social Networks dataset. This is due to the relatively simpler nature of the Small Molecules dataset, which has fewer and simpler node connections, leading to a greater emphasis on the overall structural information over node features. Consequently, SC demonstrates superior performance by leveraging structural properties, while NP, which focuses on node features, falls behind. Conversely, on the more complex Social Networks dataset, NP excels at extracting and utilizing node features, thus outperforming SC.

In addition to presenting the model test results, we also depict the training process of the three different FedGCF versions, as shown in Fig. \ref{fig:ablation_acc}. Overall, FedGCF consistently outperforms the other three versions. Specifically, on the Small Molecules dataset, to achieve a model test accuracy of 72.5\%, FedGCF only requires 49 rounds, while SC, NP, and EF require 121, 168, and 79 rounds, respectively, providing advantages of 147\%, 243\%, and 61\% in the training speed. 
Furthermore, we present the model communication cost for FedGCF, SC, NP, and EF when achieving the same model test accuracy, as shown in Fig. \ref{fig:ablation_communication}. 
Overall, FedGCF requires less communication cost to reach the target model test accuracy on both datasets. 
For instance, on the Social Networks dataset, FedGCF only needs 167MB to achieve 60\% model test accuracy, while SC, NP, and EF require 985MB, 895MB and 470MB, respectively. 
These results indicate that the components in our proposed framework can effectively extract structural properties to ensure the model understands the overall structure of the graph data and node features for complex node connections in the graph. 
Adaptive graph characteristics fusion allows for a more flexible fusion of these two types of characteristics according to the current system environment. These results underscore the necessity and importance of these three components.


\section{CONCLUSION}\label{sec:conclusion}


In this work, we propose FedGCF, a novel FGL framework specifically designed to extract structural properties and node features from graph data, fusing these two types of characteristics with an optimal combination ratio to enhance model training performance. FedGCF employs a learning-driven algorithm that adaptively adjusts the combination ratio based on the system environment and training dynamics. This adaptive adjustment mechanism enables FedGCF to optimize the fusion of structural properties and node features, allowing it to better align with the diverse demands of graph characteristics across different graph scenarios, thereby improving both model training performance and generalization capabilities. The extensive experimental results demonstrate that FedGCF significantly outperforms the existing baselines.



\bibliographystyle{IEEEtran}
\bibliography{papercontent/refs}

\begin{thebibliography}{10}
\providecommand{\url}[1]{#1}
\csname url@samestyle\endcsname
\providecommand{\newblock}{\relax}
\providecommand{\bibinfo}[2]{#2}
\providecommand{\BIBentrySTDinterwordspacing}{\spaceskip=0pt\relax}
\providecommand{\BIBentryALTinterwordstretchfactor}{4}
\providecommand{\BIBentryALTinterwordspacing}{\spaceskip=\fontdimen2\font plus
\BIBentryALTinterwordstretchfactor\fontdimen3\font minus \fontdimen4\font\relax}
\providecommand{\BIBforeignlanguage}[2]{{%
\expandafter\ifx\csname l@#1\endcsname\relax
\typeout{** WARNING: IEEEtran.bst: No hyphenation pattern has been}%
\typeout{** loaded for the language `#1'. Using the pattern for}%
\typeout{** the default language instead.}%
\else
\language=\csname l@#1\endcsname
\fi
#2}}
\providecommand{\BIBdecl}{\relax}
\BIBdecl

\bibitem{ying2018graph}
R.~Ying, R.~He, K.~Chen, P.~Eksombatchai, W.~L. Hamilton, and J.~Leskovec, ``Graph convolutional neural networks for web-scale recommender systems,'' in \emph{Proceedings of the 24th ACM SIGKDD international conference on knowledge discovery \& data mining}, 2018, pp. 974--983.

\bibitem{mislove2007measurement}
A.~Mislove, M.~Marcon, K.~P. Gummadi, P.~Druschel, and B.~Bhattacharjee, ``Measurement and analysis of online social networks,'' in \emph{Proceedings of the 7th ACM SIGCOMM conference on Internet measurement}, 2007, pp. 29--42.

\bibitem{hamilton2017representation}
W.~L. Hamilton, R.~Ying, and J.~Leskovec, ``Representation learning on graphs: Methods and applications,'' \emph{arXiv preprint arXiv:1709.05584}, 2017.

\bibitem{zhang2019heterogeneous}
C.~Zhang, D.~Song, C.~Huang, A.~Swami, and N.~V. Chawla, ``Heterogeneous graph neural network,'' in \emph{Proceedings of the 25th ACM SIGKDD international conference on knowledge discovery \& data mining}, 2019, pp. 793--803.

\bibitem{wu2019simplifying}
F.~Wu, A.~Souza, T.~Zhang, C.~Fifty, T.~Yu, and K.~Weinberger, ``Simplifying graph convolutional networks,'' in \emph{International conference on machine learning}.\hskip 1em plus 0.5em minus 0.4em\relax PMLR, 2019, pp. 6861--6871.

\bibitem{li2019deepgcns}
G.~Li, M.~Muller, A.~Thabet, and B.~Ghanem, ``Deepgcns: Can gcns go as deep as cnns?'' in \emph{Proceedings of the IEEE/CVF international conference on computer vision}, 2019, pp. 9267--9276.

\bibitem{li2022federated}
Z.~Li, M.~Bilal, X.~Xu, J.~Jiang, and Y.~Cui, ``Federated learning-based cross-enterprise recommendation with graph neural networks,'' \emph{IEEE Transactions on Industrial Informatics}, vol.~19, no.~1, pp. 673--682, 2022.

\bibitem{yao2024fedgcn}
Y.~Yao, W.~Jin, S.~Ravi, and C.~Joe-Wong, ``Fedgcn: Convergence-communication tradeoffs in federated training of graph convolutional networks,'' \emph{Advances in neural information processing systems}, vol.~36, 2024.

\bibitem{zhang2023glasu}
X.~Zhang, M.~Hong, and J.~Chen, ``Glasu: A communication-efficient algorithm for federated learning with vertically distributed graph data,'' \emph{arXiv preprint arXiv:2303.09531}, 2023.

\bibitem{wu2021fedgnn}
C.~Wu, F.~Wu, Y.~Cao, Y.~Huang, and X.~Xie, ``Fedgnn: Federated graph neural network for privacy-preserving recommendation,'' \emph{arXiv preprint arXiv:2102.04925}, 2021.

\bibitem{wang2022enhancing}
L.~Wang, Y.~Xu, H.~Xu, J.~Liu, Z.~Wang, and L.~Huang, ``Enhancing federated learning with in-cloud unlabeled data,'' in \emph{2022 IEEE 38th International Conference on Data Engineering (ICDE)}.\hskip 1em plus 0.5em minus 0.4em\relax IEEE, 2022, pp. 136--149.

\bibitem{jiang2022fedmp}
Z.~Jiang, Y.~Xu, H.~Xu, Z.~Wang, C.~Qiao, and Y.~Zhao, ``Fedmp: Federated learning through adaptive model pruning in heterogeneous edge computing,'' in \emph{2022 IEEE 38th International Conference on Data Engineering (ICDE)}.\hskip 1em plus 0.5em minus 0.4em\relax IEEE, 2022, pp. 767--779.

\bibitem{liu2020federated}
F.~Liu, X.~Wu, S.~Ge, W.~Fan, and Y.~Zou, ``Federated learning for vision-and-language grounding problems,'' in \emph{Proceedings of the AAAI conference on artificial intelligence}, vol.~34, no.~07, 2020, pp. 11\,572--11\,579.

\bibitem{yan2024peaches}
J.~Yan, J.~Liu, H.~Xu, Z.~Wang, and C.~Qiao, ``Peaches: Personalized federated learning with neural architecture search in edge computing,'' \emph{IEEE Transactions on Mobile Computing}, 2024.

\bibitem{newman2003structure}
M.~E. Newman, ``The structure and function of complex networks,'' \emph{SIAM review}, vol.~45, no.~2, pp. 167--256, 2003.

\bibitem{clayden2012organic}
J.~Clayden, N.~Greeves, and S.~Warren, \emph{Organic chemistry}.\hskip 1em plus 0.5em minus 0.4em\relax Oxford University Press, USA, 2012.

\bibitem{marin2011social}
A.~Marin and B.~Wellman, ``Social network analysis: An introduction,'' \emph{The SAGE handbook of social network analysis}, pp. 11--25, 2011.

\bibitem{gao2024towards}
X.~Gao, J.~Liu, H.~Xu, Q.~Ma, and L.~Wang, ``Towards communication-efficient federated graph learning: An adaptive client selection perspective,'' in \emph{2024 IEEE/ACM 32nd International Symposium on Quality of Service (IWQoS)}.\hskip 1em plus 0.5em minus 0.4em\relax IEEE, 2024, pp. 1--10.

\bibitem{zhao2018federated}
Y.~Zhao, M.~Li, L.~Lai, N.~Suda, D.~Civin, and V.~Chandra, ``Federated learning with non-iid data,'' \emph{arXiv preprint arXiv:1806.00582}, 2018.

\bibitem{wu2023non}
J.~Wu, J.~He, and E.~Ainsworth, ``Non-iid transfer learning on graphs,'' in \emph{Proceedings of the AAAI Conference on Artificial Intelligence}, vol.~37, no.~9, 2023, pp. 10\,342--10\,350.

\bibitem{zhang2022fine}
L.~Zhang, L.~Shen, L.~Ding, D.~Tao, and L.-Y. Duan, ``Fine-tuning global model via data-free knowledge distillation for non-iid federated learning,'' in \emph{Proceedings of the IEEE/CVF conference on computer vision and pattern recognition}, 2022, pp. 10\,174--10\,183.

\bibitem{li2019convergence}
X.~Li, K.~Huang, W.~Yang, S.~Wang, and Z.~Zhang, ``On the convergence of fedavg on non-iid data,'' \emph{arXiv preprint arXiv:1907.02189}, 2019.

\bibitem{xie2021federated}
H.~Xie, J.~Ma, L.~Xiong, and C.~Yang, ``Federated graph classification over non-iid graphs,'' \emph{Advances in neural information processing systems}, vol.~34, pp. 18\,839--18\,852, 2021.

\bibitem{tan2023federated}
Y.~Tan, Y.~Liu, G.~Long, J.~Jiang, Q.~Lu, and C.~Zhang, ``Federated learning on non-iid graphs via structural knowledge sharing,'' in \emph{Proceedings of the AAAI conference on artificial intelligence}, vol.~37, no.~8, 2023, pp. 9953--9961.

\bibitem{huang2024federated}
W.~Huang, G.~Wan, M.~Ye, and B.~Du, ``Federated graph semantic and structural learning,'' \emph{arXiv preprint arXiv:2406.18937}, 2024.

\bibitem{he2021fedgraphnn}
C.~He, K.~Balasubramanian, E.~Ceyani, C.~Yang, H.~Xie, L.~Sun, L.~He, L.~Yang, S.~Y. Philip, Y.~Rong \emph{et~al.}, ``Fedgraphnn: A federated learning benchmark system for graph neural networks,'' in \emph{ICLR 2021 Workshop on Distributed and Private Machine Learning (DPML)}, 2021.

\bibitem{kipf2016semi}
T.~N. Kipf and M.~Welling, ``Semi-supervised classification with graph convolutional networks,'' \emph{arXiv preprint arXiv:1609.02907}, 2016.

\bibitem{hamilton2017inductive}
W.~Hamilton, Z.~Ying, and J.~Leskovec, ``Inductive representation learning on large graphs,'' \emph{Advances in neural information processing systems}, vol.~30, 2017.

\bibitem{mcmahan2017communication}
B.~McMahan, E.~Moore, D.~Ramage, S.~Hampson, and B.~A. y~Arcas, ``Communication-efficient learning of deep networks from decentralized data,'' in \emph{Artificial intelligence and statistics}.\hskip 1em plus 0.5em minus 0.4em\relax PMLR, 2017, pp. 1273--1282.

\bibitem{shun2013ligra}
J.~Shun and G.~E. Blelloch, ``Ligra: a lightweight graph processing framework for shared memory,'' in \emph{Proceedings of the 18th ACM SIGPLAN symposium on Principles and practice of parallel programming}, 2013, pp. 135--146.

\bibitem{salihoglu2014optimizing}
S.~Salihoglu and J.~Widom, ``Optimizing graph algorithms on pregel-like systems,'' 2014.

\bibitem{merrill2012scalable}
D.~Merrill, M.~Garland, and A.~Grimshaw, ``Scalable gpu graph traversal,'' \emph{ACM Sigplan Notices}, vol.~47, no.~8, pp. 117--128, 2012.

\bibitem{gilmer2017neural}
J.~Gilmer, S.~S. Schoenholz, P.~F. Riley, O.~Vinyals, and G.~E. Dahl, ``Neural message passing for quantum chemistry,'' in \emph{International conference on machine learning}.\hskip 1em plus 0.5em minus 0.4em\relax PMLR, 2017, pp. 1263--1272.

\bibitem{zhou2020graph}
J.~Zhou, G.~Cui, S.~Hu, Z.~Zhang, C.~Yang, Z.~Liu, L.~Wang, C.~Li, and M.~Sun, ``Graph neural networks: A review of methods and applications,'' \emph{AI open}, vol.~1, pp. 57--81, 2020.

\bibitem{hsieh2020non}
K.~Hsieh, A.~Phanishayee, O.~Mutlu, and P.~Gibbons, ``The non-iid data quagmire of decentralized machine learning,'' in \emph{International Conference on Machine Learning}.\hskip 1em plus 0.5em minus 0.4em\relax PMLR, 2020, pp. 4387--4398.

\bibitem{bojchevski2018netgan}
A.~Bojchevski, O.~Shchur, D.~Z{\"u}gner, and S.~G{\"u}nnemann, ``Netgan: Generating graphs via random walks,'' in \emph{International conference on machine learning}.\hskip 1em plus 0.5em minus 0.4em\relax PMLR, 2018, pp. 610--619.

\bibitem{reddi2020adaptive}
S.~Reddi, Z.~Charles, M.~Zaheer, Z.~Garrett, K.~Rush, J.~Kone{\v{c}}n{\`y}, S.~Kumar, and H.~B. McMahan, ``Adaptive federated optimization,'' \emph{arXiv preprint arXiv:2003.00295}, 2020.

\bibitem{xia2019random}
F.~Xia, J.~Liu, H.~Nie, Y.~Fu, L.~Wan, and X.~Kong, ``Random walks: A review of algorithms and applications,'' \emph{IEEE Transactions on Emerging Topics in Computational Intelligence}, vol.~4, no.~2, pp. 95--107, 2019.

\bibitem{tong2006fast}
H.~Tong, C.~Faloutsos, and J.-Y. Pan, ``Fast random walk with restart and its applications,'' in \emph{Sixth international conference on data mining (ICDM'06)}.\hskip 1em plus 0.5em minus 0.4em\relax IEEE, 2006, pp. 613--622.

\bibitem{albalahi2022vertex}
A.~M. Albalahi, I.~Z. Milovanovic, Z.~Raza, A.~Ali, and A.~E. Hamza, ``On the vertex-degree-function indices of connected (n, m)-graphs of maximum degree at most four,'' \emph{arXiv preprint arXiv:2207.00353}, 2022.

\bibitem{dann2017unifying}
C.~Dann, T.~Lattimore, and E.~Brunskill, ``Unifying pac and regret: Uniform pac bounds for episodic reinforcement learning,'' \emph{Advances in Neural Information Processing Systems}, vol.~30, 2017.

\bibitem{schulman2015trust}
J.~Schulman, ``Trust region policy optimization,'' \emph{arXiv preprint arXiv:1502.05477}, 2015.

\bibitem{schulman2017proximal}
J.~Schulman, F.~Wolski, P.~Dhariwal, A.~Radford, and O.~Klimov, ``Proximal policy optimization algorithms,'' \emph{arXiv preprint arXiv:1707.06347}, 2017.

\bibitem{auer2002finite}
P.~Auer, ``Finite-time analysis of the multiarmed bandit problem,'' 2002.

\bibitem{zhou2015survey}
L.~Zhou, ``A survey on contextual multi-armed bandits,'' \emph{arXiv preprint arXiv:1508.03326}, 2015.

\bibitem{li2010contextual}
L.~Li, W.~Chu, J.~Langford, and R.~E. Schapire, ``A contextual-bandit approach to personalized news article recommendation,'' in \emph{Proceedings of the 19th international conference on World wide web}, 2010, pp. 661--670.

\bibitem{xu2018powerful}
K.~Xu, W.~Hu, J.~Leskovec, and S.~Jegelka, ``How powerful are graph neural networks?'' \emph{arXiv preprint arXiv:1810.00826}, 2018.

\bibitem{li2020federated}
T.~Li, A.~K. Sahu, M.~Zaheer, M.~Sanjabi, A.~Talwalkar, and V.~Smith, ``Federated optimization in heterogeneous networks,'' \emph{Proceedings of Machine learning and systems}, vol.~2, pp. 429--450, 2020.

\bibitem{kingma2014adam}
D.~P. Kingma, ``Adam: A method for stochastic optimization,'' \emph{arXiv preprint arXiv:1412.6980}, 2014.

\bibitem{reddi2019convergence}
S.~J. Reddi, S.~Kale, and S.~Kumar, ``On the convergence of adam and beyond,'' \emph{arXiv preprint arXiv:1904.09237}, 2019.

\end{thebibliography}

\end{document}